\documentclass[12pt]{article}
\pdfoutput=1
\usepackage{times}
\usepackage{natbib}
 \bibpunct{(}{)}{;}{a}{,}{,}
\usepackage{graphicx}
\usepackage{amsmath,amssymb,amsthm}
\usepackage{bm}
\usepackage{rotating}
\usepackage[lined,ruled]{algorithm2e}
\usepackage{multirow}
\usepackage{rotating}
\usepackage{multicol}
\usepackage{titlesec}
\usepackage{latexsym}
\usepackage[pdftex,colorlinks=true,linkcolor=blue,citecolor=blue,urlcolor=blue,bookmarks=false,pdfpagemode=None]{hyperref}
\usepackage{url}
\usepackage{verbatim}
\usepackage{fancyhdr}
\usepackage{setspace}
\usepackage{paralist}
\usepackage{boxedminipage}
\usepackage{lineno}
\usepackage[top=1in, bottom=1in, left=1in, right=1in]{geometry}
\usepackage{lscape}
\usepackage{dcolumn}

\usepackage[usenames,dvipsnames]{color}
\newcommand{\rev}[1]{\color{black}#1 \color{black}}

\newcolumntype{.}{D{.}{.}{-1}}

\newcommand{\bv}{\begin{array}}

\def\abovestrut#1{\rule[0in]{0in}{#1}\ignorespaces}
\def\belowstrut#1{\rule[-#1]{0in}{#1}\ignorespaces}
\def\abovespace{\abovestrut{0.20in}}

\def\belowspace{\belowstrut{0.10in}}

\newenvironment{packed_itemize}{
\begin{itemize}
  \linespread{0.25}
  \setlength{\itemsep}{2pt}
  \setlength{\parskip}{0pt}
  \setlength{\parsep}{0pt}
}{\end{itemize}}

\begin{document}

\title{A Poisson convolution model for characterizing topical content with word frequency and exclusivity
}

\author{
 Edoardo M. Airoldi,~ Jonathan M. Bischof\\
 Department of Statistics\\
 Harvard University, Cambridge, MA 02138, USA}
\date{}

\maketitle


\newpage

\begin{abstract}
An ongoing challenge in the analysis of document collections is how to summarize content in terms of a set of inferred \textit{themes} that can be interpreted substantively in terms of topics.
The current practice of parametrizing the themes in terms of most frequent words limits interpretability by ignoring the differential use of words across topics. We argue that words that are both common and exclusive to a theme are more effective at characterizing topical content. 
We consider a setting where professional editors have annotated documents to a collection of topic categories, organized into a tree, in which leaf-nodes correspond to the most specific topics. Each document is annotated to multiple categories, at different levels of the tree. \rev{We introduce a hierarchical Poisson convolution model to analyze annotated documents in this setting. 
The model leverages the structure among categories defined by professional editors to infer a clear semantic description for each topic in terms of words that are both frequent and exclusive. 
We carry out a large randomized experiment on Amazon Turk to demonstrate that 
 topic summaries based on the FREX score are more interpretable than currently established frequency based summaries, and that the proposed model produces more efficient estimates of exclusivity than with currently models.
We also develop a parallelized Hamiltonian Monte Carlo sampler that allows the inference to scale to millions of documents.}

\vfill
\end{abstract}

\newpage
\singlespacing
\small
\tableofcontents
\normalsize
\doublespacing

\newpage


\section{Introduction}
\label{sec-intro}





A recurring challenge in multivariate statistics is how to construct interpretable low-dimensional summaries of high-dimensional data. 
Historically, simple models based on correlation matrices, such as principal component analysis \citep{Joll:1986} and canonical correlation analysis \citep{hote:1936}, have proven to be effective tools for data reduction. More recently, multilevel models have become a flexible and powerful tool for finding latent structure in high dimensional data \citep{mclachlanpeel2000,sohnxing2009,bleietal03,airoldietal08}.
%
%
However, while  interpretable statistical summaries are highly valued in applications, dimensionality reduction models are rarely optimized to aid qualitative discovery; there is no guarantee that the optimal low-dimensional projections will be understandable in terms of quantities of scientific interest that can help practitioners make decisions. 
Instead, we design a model with scientific estimands of interest in mind to achieve an optimal balance of interpretability and dimensionality reduction. 


We consider a setting in which we observe two sets of categorical data for each unit of observation: $\bm{w}_{1:V}$, which live in a high-dimensional space, and $\bm{l}_{1:K}$, which live in a structured low-dimensional space and provide a direct link to information of scientific interest about the sampling units. The goal of the analysis is two fold. First, we desire to develop a joint model for the observations $\bm{Y} \equiv \{\bm{W}_{D\times{}V}, \bm{L}_{D\times{}K}\}$ that can be used to project the data onto a low-dimensional parameter space $\bm{\Theta}$ in which interpretability is maintained by mapping categories in $\mathcal{L}$ to directions in $\bm{\Theta}$. Second, we would like the mapping from the original space to the low-dimensional projection to be scientifically interesting so that statistical insights about $\bm{\Theta}$ can be understood in terms of the original inputs, $\bm{w}_{1:V}$, in a way that guides future research.


In the application to text analysis that motivates this work, $\bm{w}_{1:N}$ are the raw word counts observed in each document and $\bm{l}_{1:K}$ are a set of labels created by professional editors that are indicative of topical content. Specifically, the words are represented as an unordered vector of counts, with the length of the vector corresponding to the size of a known dictionary. The labels are organized in a tree-structured ontology, from the most generic topic at the root of the tree to the most specific topic at the leaves. 
Each news article may be annotated with more than one label, at the editors' discretion. The number of labels is given by the size of the ontology and typically ranges from tens to hundreds of categories. In this context, the inferential challenge is to discover a low dimensional representation of topical content, $\bm{\Theta}$, that aligns with the coarse labels provided by editors while at the same time providing a mapping between the textual content and directions in $\bm{\Theta}$ in a way that formalizes and enhances our understanding of how low dimensional structure is expressed the space of observed words.


Recent approaches to this problem in the machine learning literature have taken a Bayesian hierarchical approach to this task by viewing a document's content as arising from a mixture of component distributions, commonly referred to as ``topics" as they often capture thematic structure \citep{blei:2012}. As the component distributions are almost exclusively parameterized as multinomial distributions over words in the vocabulary, the loading of words onto topics is characterized in terms of the relative frequency of within-component usage. While relative frequency has proven to be a useful mapping of topical content onto words, recent work has documented a growing list of interpretability issues with frequency-based summaries: they are often dominated by contentless ``stop" words \citep{wallachetal09}, sometimes appear incoherent or redundant \citep{mimnoetal09,changetal09}, and typically require post hoc modification to meet human expectations \citep{huetal11}. 
Instead, we propose a new mapping for topical content that incorporates how words are used differentially across topics.  If a word is common in a topic, it is also important to know whether it is common in many topics or relatively exclusive to the topic in question. Both of these summary statistics are informative: nonexclusive words are less likely to carry topic-specific content, while infrequent words occur too rarely to form the semantic core of a topic. We therefore look for the most frequent words in the corpus that are also likely to have been generated from the topic of interest to summarize its content. 
In this approach we borrow ideas from the statistical literature, in which models of differential word usage have been leveraged for  analyzing writing styles in a supervised setting \citep{Most:Wall:1984,Airo:Ande:Fien:Skin:2006}, and combine them with ideas from the machine learning literature, in which latent variable and mixture models based on frequent word usage have been used to infer structure that often captures topical content \citep{mccallumetal98, bleietal03,canny04}.

\rev{From a statistical perspective, 
models based on topic-specific distributions over the vocabulary \cite{bleietal03} often fail to produce stable estimates of differential word usage since they only model the relative frequency of words within topics. Results in \citet{eisensteinetal11} suggest that such a popular parametrization leads to an amplification of the estimated differential word usage rates, especially for rare words, arguably because of the lack of mechanisms to regularize word rates across topics. The trade-off between these two orthogonal regularization strategies (over words within a topic versus over the same word across topics) has been explored in the literature \citep{Most:Wall:1964,Most:Wall:1984,canny04,Airo:Fien:Xing:2006}. To tackle this issue, we introduce the generative framework of Hierarchical Poisson Convolution (HPC) that parameterizes topic-specific word counts as unnormalized count variates whose rates can be regularized across topics as well as within them, leading to stable inference of both word frequency and exclusivity, as we show in Section \ref{sec:mturk}.} HPC can be seen as a fully generative extension of Sparse Topic Coding \citep{zhuxing11} that emphasizes regularization and interpretability rather than exact sparsity. Additionally, HPC leverages hierarchical systems of topic categories created by professional editors in collections such as \emph{Reuters}, \emph{New York Times}, \emph{Wikipedia}, and \emph{Encyclopedia Britannica} to make focused comparisons of differential use between neighboring topics on the tree and build a sophisticated joint model for topic memberships and labels in the documents. By conditioning on a known hierarchy, we avoid the complicated task of inferring hierarchical structure \citep{bleihier03,mimnoetal07,adams-ghahramani-jordan-2010}. We introduce a parallelized Hamiltonian Monte Carlo (HMC) estimation strategy that makes full Bayesian inference efficient and scalable.

The proposed model is designed to infer an interpretable description of human-generated labels, thus we restrict the topic components to have a one-to-one correspondence with the human-generated labels, as in Labeled LDA \citep{ramageetal09}. This \emph{descriptive} link between the labels and topics differs from the \emph{predictive} link used in Supervised LDA \citep{bleimcauliffe07,perotteetal12}, where topics are learned as an optimal covariate space to predict an observed document label or response variable. The more restrictive descriptive link can be expected to limit predictive power, but leads to summaries directly associated with individual labels. We then infer a description of these labels in terms of words that are both frequent and exclusive. We anticipate that learning a concise semantic description for any collection of topics implicitly defined by professional editors is the first step toward the semi-automated creation of domain-specific topic ontologies. Domain-specific topic ontologies may be useful for evaluating the semantic content of \textit{inferred} topics, or for predicting the semantic content of new social media, including Twitter messages and Facebook wall-posts.

\section{Hierarchical Poisson Convolution}\label{sec-model}

The Hierarchical Poisson Convolution (HPC) model is a data generating process for document collections whose topics are organized in a hierarchy, and whose topic labels are observed. We refer to the structure among topics interchangeably as a \emph{hierarchy} or \emph{tree} since we assume that each topic has exactly one parent and that no cyclical parental relations are allowed. Each document $d\in\{1,\ldots,D\}$ is a record of counts $w_{fd}$ for every feature in the vocabulary, $f\in\{1,\ldots,V\}$. The length of the document is given by $L_{d}$, which we normalize by the average document length $L$ to get $l_{d}\equiv{}\frac{1}{L}L_{d}$ \citep{Most:Wall:1984,Airo:Ande:Fien:Skin:2006}.  Documents have unrestricted membership to any combination of topics $k\in\{1,\ldots,K\}$ represented by a vector of labels $\bm{I}_{d}$ where $I_{dk}\equiv{}I\{\textrm{doc $d$ belongs to topic $k$}\}$.



\begin{figure}[t!]
\caption{Graphical representation of Hierarchical Poisson Convolution (left) and detail on tree plate (right)}
\label{fig-model}
\begin{center}
\includegraphics[width=0.8\textwidth]{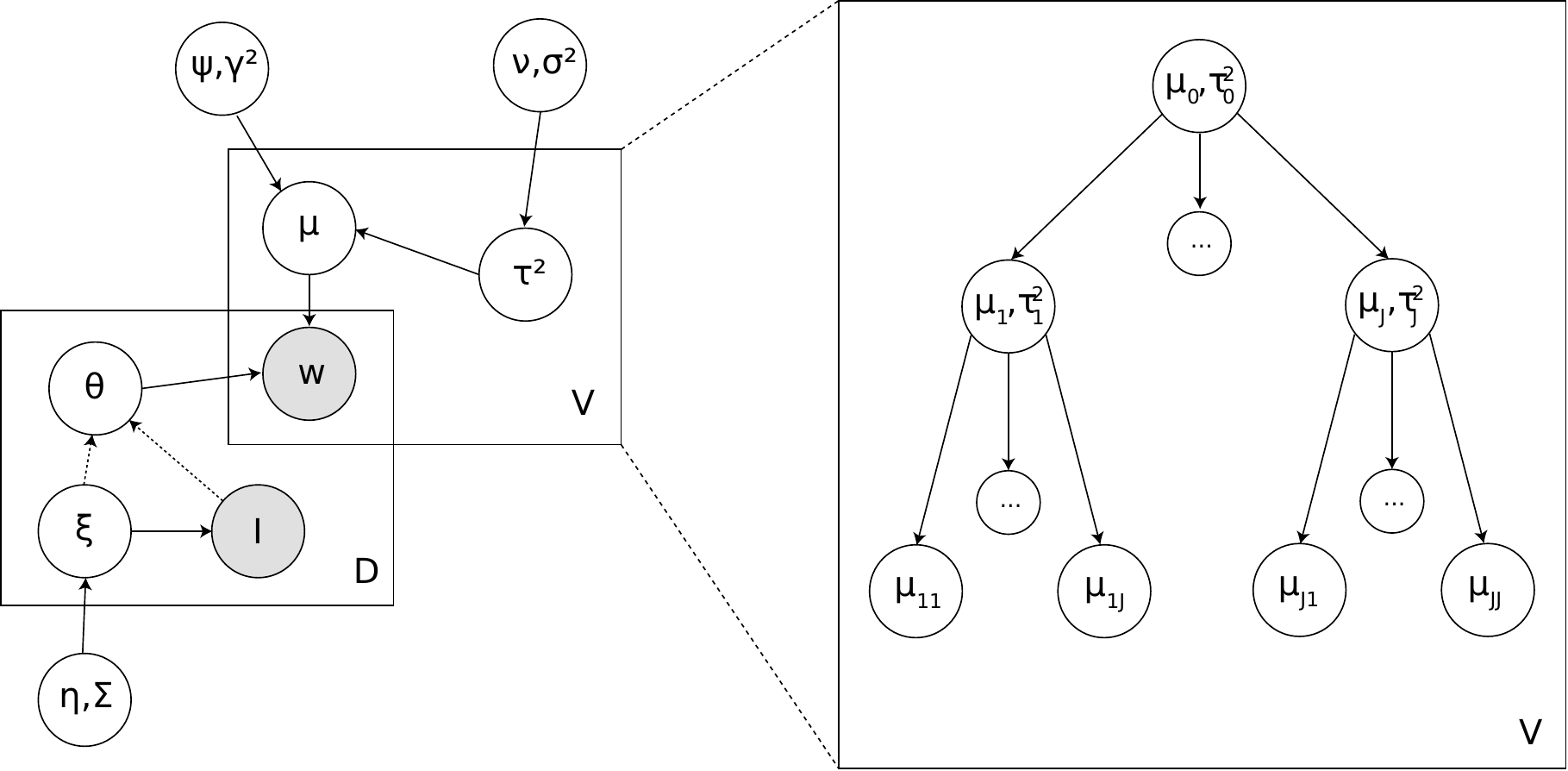}
\end{center}
\end{figure}

\subsection{Modeling word usage rates on the hierarchy}

The HPC model leverages the known topic hierarchy by assuming that words are used similarly in neighboring topics. 
\rev{Let $\beta_{f,k}$ be the occurrence rate for word $f$ in topic $k$, and define $\mu_{f,k} \equiv \beta_{f,k}$ for convenience of notation.}
Specifically, the log rate for a word across topics follows a Gaussian diffusion down the tree. Consider the topic hierarchy presented in the right panel of Figure \ref{fig-model}. At the top level, $\mu_{f,0}$ represents the log rate for feature $f$ overall in the corpus. The log rates $\mu_{f,1},\ldots,\mu_{f,J}$ for first level topics 
are then drawn from a Gaussian centered around the corpus rate with dispersion controlled by the variance parameter $\tau^{2}_{f,0}$. From first level topics,
we then draw the log rates for the second level topics
 from another Gaussian centered around their mean $\mu_{f,j}$ and with variance $\tau^{2}_{f,j}$. This process is continued down the tree, with each parent node having a separate variance parameter to control the dispersion of its children. 
 
The variance parameters $\tau^{2}_{fp}$ directly control the local differential expression in a branch of the tree. Words with high variance parameters can have rates in the child topics that differ greatly from the parent topic $p$, allowing the child rates to diverge. Words with low variance parameters will have rates close to the parent and so will be expressed similarly among the children. If we learn a population distribution for the $\tau^{2}_{fp}$ that has low mean and variance, it is equivalent to saying that most features are expressed similarly across topics \textit{a priori} and that we would need a preponderance of evidence to believe otherwise.

\rev{Because of the hierarchy on the rates of word occurrence, the typical equivalence between an array of Poisson distributions for topic-specific word counts and a Poisson distribution for the total  counts combined with a Multinomial distribution to allocate counts across topics \citep[e.g.,see][]{canny04,Buntin:2006fk,Airo:Fien:Xing:2006,eisensteinetal11} no longer holds.}


\subsection{Modeling the topic membership of documents}
Documents in the HPC model can contain content from any of the $K$ topics in the hierarchy at varying proportions, with the exact allocation given by the vector $\bm{\theta}_{d}$ on the $K-1$ simplex. The model assumes that the count for word $f$ contributed by each topic follows a Poisson distribution whose rate is moderated by the document's length and membership to the topic; that is, $w_{fdk} \sim \text{Pois}(l_{d}\theta_{dk}\beta_{fk})$. The only data we observe is the total word count $w_{fd} \equiv \sum_{k=1}^{K}w_{fdk}$, but the infinite divisibility property of the Poisson distribution gives us that $w_{fd}\sim\textrm{Pois}(l_{d}\bm{\theta}_{d}^{T}\bm{\beta}_{f})$. These draws are done for every word in the vocabulary (using the same $\bm{\theta}_{d}$) to get the content of the document.\footnote{This is where the model's name arises: the observed feature count in each document is the convolution of (unobserved) topic-specific Poisson variates.}


In labeled document collections, human coders provide an extra piece of information for each document, $\bm{I}_{d}$, that indicates the set of topics that contributed its content. As a result, we know $\theta_{dk}=0$ for all topics $k$ where $I_{dk}=0$, and only have to determine how content is allocated between the set of active topics. \rev{The data generating process in Table \ref{table-genprocess} leads to well defined topic proportions $\bm{\theta}_d$ when at least one element of $\bm{I}_d$ is positive. This  constraint is a non-issue, however, since we are implicitly conditioning on $I_{dk}>1$ for some $k$ when fitting the model to data.}

The HPC model assumes that these two sources of information for a document are not generated independently. A document should not have a high probability of being labeled to a topic from which it receives little content and vice versa. Instead, the model posits a latent $K$-dimensional topic affinity vector $\bm{\xi}_{d}\sim{}\mathcal{N}(\bm{\eta},\bm{\Sigma})$ that expresses how strongly the document is associated with each topic. The topic memberships and labels of the document are different manifestations of this affinity. Specifically, each $\xi_{dk}$ is the log odds that topic label $k$ is active in the document, with $I_{dk}\sim \text{Bernoulli}(\text{logit}^{-1}(\xi_{dk}))$. Conditional on the labels, the topic memberships are the relative sizes of the document's affinity for the active topics and zero for inactive topics: $\theta_{dk}\equiv{}e^{\xi_{dk}}I_{dk}/\sum_{j=1}^{K}e^{\xi_{dj}}I_{dj}$. Restricting each document's membership vectors to the labeled topics is a natural and efficient way to generate sparsity in the mixing parameters, stabilizing inference and reducing the computational burden of posterior simulation.


\begin{table}[b!]
\small
\setlength{\tabcolsep}{20pt}
\caption{Generative process for Hierarchical Poisson Convolution}
\label{table-genprocess}
\begin{center}
\begin{tabular}{p{0.19\textwidth}p{0.63\textwidth}}
\bf Step  & \bf Generative process
\\ \hline \\
Tree parameters\- & For feature $f \in \{1,\ldots,V\}$:
\begin{packed_itemize}
 \item Draw $\mu_{f,0} \sim \mathcal{N}(\psi,\gamma^{2})$
 \item Draw $\tau_{f,0}^{2} \sim \textrm{Scaled Inv-}\chi^{2}(\nu,\sigma^{2})$
  \item For $j \in \{1,\ldots,J\}$ (first level of hierarchy):
\begin{packed_itemize}
 \item Draw $\mu_{f,j} \sim \mathcal{N}(\mu_{f,0},\tau_{f,0}^{2})$
 \item Draw $\tau_{f,j}^{2} \sim \textrm{Scaled Inv-}\chi^{2}(\nu,\sigma^{2})$
\end{packed_itemize}
 \item For $j \in \{1,\ldots,J\}$ (terminal level of hierarchy):
\begin{packed_itemize}
 \item Draw $\mu_{f,j1},\ldots,\mu_{f,jJ} \sim \mathcal{N}(\mu_{f,j},\tau_{f,j}^{2})$
\end{packed_itemize}
 \item Define $\beta_{f,k}\equiv{}e^{\mu_{f,k}} \text{ for } k\in\{1,\ldots,K\}$
\end{packed_itemize}
\\

Topic membership parameters\- &
For document $d\in\{1,\ldots,D\}$:
\begin{packed_itemize}
 \item Draw $\bm{\xi}_{d}\sim{}\mathcal{N}(\bm{\eta},\bm{\Sigma}=\lambda^2\bm{I}_{K})$
 \item For topic $k\in\{1,\ldots,K\}$:
\begin{packed_itemize}
 \item Define $p_{dk}\equiv{}1/(1+e^{-\xi_{dk}})$
 \item Draw $I_{dk}\sim{}\textrm{Bernoulli}(p_{dk})$
 \item Define $\theta_{dk}(\bm{I}_{d},\bm{\xi}_{d})\equiv{}e^{\xi_{dk}}I_{dk}/\sum_{j=1}^{K}e^{\xi_{dj}}I_{dj},$\newline
 \rev{where $I_{dj}>0 \hbox{ for some } j$}
\end{packed_itemize}
\end{packed_itemize} \\

Data generation\- &
For document $d\in\{1,\ldots,D\}$:
\begin{packed_itemize}
 \item Draw normalized document length \rev{$l_{d}L\sim{}\textrm{Pois}(\upsilon)$}
 \item For every topic $k$ and feature $f$:
\begin{packed_itemize}
 \item Draw count $w_{fdk}\sim\textrm{Pois}(l_{d}\bm{\theta}_{d}^{T}\bm{\beta}_{f})$ 
\end{packed_itemize}
 \item Define $w_{fd} \equiv \sum_{k=1}^{K}w_{fdk}$ \;\;(observed data)
\end{packed_itemize} \\
\hline
\end{tabular}
\end{center}
\vspace{-0.2in}
\end{table}

We outline the generative process in full detail in Table \ref{table-genprocess}, which can be summarized in three steps.  First, a set of rate and variance parameters are drawn for each feature in the vocabulary. Second, a topic affinity vector is drawn for each document in the corpus, which generate topic labels. Finally, both sets of parameters are then used to generate the words in each document. For simplicity of presentation we assume that each non-terminal node has $J$ children and that the tree has only two levels below the corpus level, but the model can accommodate any tree structure.

\subsection{Estimands}\label{sec:estimands}

In order to measure topical semantic content, we consider the topic-specific frequency and exclusivity of each word in the vocabulary. These quantities form a two-dimensional summary of each word's relation to a topic of interest, with higher scores in both being positively related to topic specific content. Additionally, we develop a univariate summary of semantic content that can be used to rank words in terms of their semantic content. These estimands are simple functions of the rate parameters of HPC; the distribution of the documents' topic memberships is a nuisance parameter needed to disambiguate the content of a document between its labeled topics.

A word's topic-specific frequency, $\beta_{fk}\equiv\exp\mu_{fk}$, is directly parameterized in the model and is regularized across words (via hyperparameters $\psi$ and $\gamma^2$) and across topics. A word's exclusivity to a topic, $\phi_{f,k}$, is its usage rate relative to a set of comparison topics $\mathcal{S}$:  $\phi_{f,k}=\beta_{f,k}/\sum_{j\in\mathcal{S}}\beta_{f,j}$. A topic's siblings are a natural choice for a comparison set to see which words are overexpressed in the topic compared to a set of similar topics. While not directly modeled in HPC, the exclusivity parameters are also regularized by the $\tau^{2}_{fp}$, since if the child rates are forced to be similar then the $\phi_{f,k}$ will be pushed toward a baseline value of $1/|\mathcal{S}|$. We explore the regularization structure of the model empirically in Section \ref{sec-res}. 
\rev{While the set $\mathcal{S}$ can be taken to be the set of all topics, in the analysis we focus on the arguably most difficult task of thematically distinguishing between pairs of closely related topics. Success in this task requires the  topical summaries to be descriptive of closely related themes, while being quantitatively and qualitatively indicative of the differences.}

Since both frequency and exclusivity are important factors in determining a word's semantic content, a univariate measure of topical importance is a useful estimand for diverse tasks such as dimensionality reduction, feature selection, and content discovery. In constructing a composite measure, we do not want a high rank in one dimension to be able to compensate for a low rank in the other since frequency or exclusivity alone are not necessarily useful. We therefore adopt the harmonic mean to pull the ``average" rank toward the lower score. For word $f$ in topic $k$, we define the $FREX_{fk}$ score as the harmonic mean of the word's rank in the distribution of $\phi_{.,k}$ and $\mu_{.,k}$:
\begin{equation*}
FREX_{fk} = \left(\frac{w}{\text{ECDF}_{\phi_{.,k}}(\phi_{f,k})} + \frac{1-w}{\text{ECDF}_{\mu_{.,k}}(\mu_{f,k})}\right)^{-1}.
\end{equation*}
where $w$ is the weight for exclusivity (which we set to $0.5$ as a default) and $\text{ECDF}_{x_{.,k}}$ is the empirical CDF function applied to the values $x$ over the first index.
\vspace{-10pt}

%
%

\section{Scalable inference via parallelized HMC sampler}\label{sec-inf}

We use a Gibbs sampler to obtain the posterior expectations of the unknown rate and membership parameters (and associated hyperparameters) given the observed data. Specifically, inference is conditioned on $\bm{W}$, a $D\times{}V$ matrix of word counts, $\bm{I}$, a $D\times{}K$ matrix of topic labels, $\bm{l}$, a $D$-vector of document lengths, and $\mathcal{T}$, a tree structure for the topics.

Creating a scalable inference method is critical since the space of latent variables grows linearly in the number of words and documents, with $K(D+V)$ total unknowns. Our model offers an advantage in that the posterior consists of two groups of parameters whose conditional posterior factors given the other. On one side, the conditional posterior of the rate and variance parameters $\{\bm{\mu}_{f},\bm{\tau}_{f}^2\}_{f=1}^{V}$ factors by word given the membership parameters and the hyperparameters $\psi$, $\gamma^{2}$, $\nu$ and $\sigma^{2}$. On the other, the conditional posterior of the topic affinity parameters $\{\bm{\xi}_{d}\}_{d=1}^{D}$ factors by document given the hyperparameters $\bm{\eta}$ and $\bm{\Sigma}$ and the rate parameters $\{\bm{\mu}_{f}\}_{f=1}^{V}$.

Conditional on the hyperparameters, we are left with two blocks of draws that can be broken into $V$ or $D$ independent threads. Using parallel computing software such as Message Passing Interface (MPI), the computation time for drawing the parameters in each block is only constrained by resources required for a single draw. The total runtime need not significantly increase with the addition of more documents or words as long as the number of available cores also increases.

Both of these conditional distributions are only known up to a constant and can be high dimensional if there are many topics, making direct sampling impossible and random walk Metropolis inefficient. We are able to obtain uncorrelated draws through the use of Hamiltonian Monte Carlo (HMC) \citep{neal2011}, which leverages the posterior gradient and Hessian to find a distant point in the parameter space with high probability of acceptance. HMC works well for log densities that are unimodal and have relatively constant curvature. We give step-by-step instructions for our implementation of the algorithm in the Appendix.

After appropriate initialization, we follow a fixed Gibbs scan where the two blocks of latent variables 
are drawn in parallel from their conditional posteriors using HMC. We then draw the hyperparameters conditional on all the inputed latent variables.
\vspace{-10pt}

\subsection{Block Gibbs Sampler}

To set up the block Gibbs sampling algorithm, we derive the relavant conditional posterior distributions and explain how we sample from each. 
%
%
%
%

\subsubsection{Updating tree parameters}
In the first block, the conditional posterior of the tree parameters factors by word:
\begin{multline*}
p(\{\bm{\mu}_{f},\bm{\tau}_{f}^2\}_{f=1}^{V}|\bm{W},\bm{I},\bm{l},\psi,\gamma^{2},\nu,\sigma^{2},\{\bm{\xi}_{d}\}_{d=1}^{D},\mathcal{T})\varpropto\\
\prod_{f=1}^{V}\bigg\{\prod_{d=1}^{D}p(w_{fd}|\bm{I}_{d},l_{d},\mu_{f},\bm{\xi}_{d})\bigg\}\cdot{}p(\bm{\mu}_{f},\bm{\tau}_{f}^2|\psi,\gamma^{2},\mathcal{T},\nu,\sigma^{2}).
\end{multline*}
Given the conditional conjugacy of the variance parameters and their strong influence on the curvature of the rate parameter posterior, we sample the two groups conditional on each other to optimize HMC performance. Conditioning on the variance parameters, we can write the likelihood of the rate parameters as a Poisson regression where the documents are observations, the $\bm{\theta}_{d}(\bm{I}_{d},\bm{\xi}_{d})$ are the covariates, and the $l_{d}$ serve as exposure weights. 

\normalsize{The prior distribution of the rate parameters is a Gaussian graphical model, so \textit{a priori} the log rates for each word are jointly Gaussian with mean $\psi\bm{1}$ and precision matrix $\bm{\Lambda}(\gamma^{2},\bm{\tau}_{f}^2,\mathcal{T})$ which has non-zero entries only for topic pairs that have a direct parent-child relationship.\footnote{In practice this precision matrix can be found easily as the negative Hessian of the log prior distribution.}
The log conditional posterior is:}
\begin{multline*}
\log{}p(\bm{\mu}_{f}|\bm{W},\bm{I},\bm{l},\{\bm{\tau}_{f}^2\}_{f=1}^{V},\psi,\gamma^{2},\nu,\sigma^{2},\{\bm{\xi}_{d}\}_{d=1}^{D},\mathcal{T})=\\
-\sum_{d=1}^{D}l_{d}\bm{\theta}_{d}^{T}\bm{\beta}_{f} + \sum_{d=1}^{D}w_{fd}\log{}(\bm{\theta}_{d}^{T}\bm{\beta}_{f}) - \frac{1}{2}(\bm{\mu}_{f}-\psi\bm{1})^{T}\bm{\Lambda}(\bm{\mu}_{f}-\psi\bm{1}).
\end{multline*}
We use HMC to sample from this unnormalized density. Note that the covariate matrix $\bm{\Theta}_{D\times{}K}$ is very sparse in most cases, so we speed computation with a sparse matrix representation. 

We know the conditional distribution of the variance parameters due to the conjugacy of the $\text{Inverse-}\chi^{2}$ prior with the normal distribution of the log rates. Specifically, if $\mathcal{C}(\mathcal{T})$ is the set of child topics of topic $k$ with cardinality $J$, then
\begin{equation*}
\tau_{fk}^2|\bm{\mu}_{f},\nu,\sigma^{2},\mathcal{T}\sim\text{Inv-}\chi^{2}\bigg(J + \nu, \frac{\nu\sigma^{2} + \sum_{j\in\mathcal{C}}(\mu_{fj}-\mu_{fk})^{2}}{J + \nu}\bigg).
\end{equation*}

\subsubsection{Updating topic affinity parameters}

In the second block, the conditional posterior of the topic affinity vectors factors by document:

\begin{multline*}
p(\{\bm{\xi}_{d}\}_{d=1}^{D}|\bm{W},\bm{I},\bm{l},\{\bm{\mu}_{f}\}_{f=1}^{V},\bm{\eta},\bm{\Sigma})\varpropto\prod_{d=1}^{D}\bigg\{\prod_{f=1}^{V}p(w_{fd}|\bm{I}_{d},l_{d},\mu_{f},\bm{\xi}_{d})\bigg\}\cdot{}p(\bm{I}_{d}|\bm{\xi}_{d})\cdot{}p(\bm{\xi}_{d}|\bm{\eta},\bm{\Sigma}).
\end{multline*}
We can again write the likelihood as a Poisson regression, now with the rates as covariates. The log conditional posterior for one document is:
\begin{multline*}
\log{}p(\bm{\xi}_{d}|\bm{W},\bm{I},\bm{l},\{\bm{\mu}_{f}\}_{f=1}^{V},\bm{\eta},\bm{\Sigma}) =\\
-l_{d}\sum_{f=1}^{V}\bm{\beta}_{f}^{T}\bm{\theta}_{d} + \sum_{f=1}^{V}w_{fd}\log{}(\bm{\beta}_{f}^{T}\bm{\theta}_{d}) - \sum_{k=1}^{K}\log(1+e^{-\xi_{dk}})\\
 - \sum_{k=1}^{K}(1-I_{dk})\xi_{dk}-\frac{1}{2}(\bm{\xi}_{d}-\bm{\eta})^{T}\bm{\Sigma}^{-1}(\bm{\xi}_{d}-\bm{\eta}).
\end{multline*}

We use HMC to sample from this unnormalized density. Here the parameter vector $\bm{\theta}_{d}$ is sparse rather than the covariate matrix $\bm{B}_{V\times{}K}$. If we remove the entries of $\bm{\theta}_{d}$ and columns of $\bm{B}$ pertaining to topics $k$ where $I_{dk}=0$, then we are left with a low dimensional regression where only the active topics are used as covariates, greatly simplifying computation.

\subsubsection{Updating corpus-level parameters}
We draw the hyperparameters after each iteration of the block update. We put flat priors on these unknowns so that we can learn their most likely values from the data. As a result, their conditional posteriors only depend on the latent variables they generate. 

The log corpus-level rates $\mu_{f,0}$ for each word follow a Gaussian distribution with mean $\psi$ and variance $\gamma^2$. The conditional distribution of these hyperparameters is available in closed form:
\begin{eqnarray*}
& \psi|\gamma^2,\{\mu_{f,0}\}_{f=1}^{V} \sim \mathcal{N}\bigg(\frac{1}{V}\sum_{f=1}^{V}\mu_{f,0},\;\;\frac{\gamma^2}{V}\bigg),\\
& \text{and }\;\;\gamma^2 | \psi, \{\mu_{f,0}\}_{f=1}^{V} \sim \textrm{Inv-}\chi^{2}\bigg(V,\;\;\frac{1}{V}\sum_{f=1}^{V}(\mu_{f,0}-\psi)^{2}\bigg).
\end{eqnarray*}

The discrimination parameters $\tau^{2}_{fk}$ independently follow an identical Scaled Inverse-$\chi^{2}$ with convolution parameter $\nu$ and scale parameter $\sigma^{2}$, while their inverse follows a $\text{Gamma}(\kappa_{\tau}=\frac{\nu}{2},\lambda_{\tau}=\frac{2}{\nu\sigma^2})$ distribution. We use HMC to sample from this unnormalized density. Specifically,
\begin{multline*}
\log{}p(\kappa_{\tau},\lambda_{\tau}|\{\bm{\tau}_{f}^2\}_{f=1}^{V},\mathcal{T}) = (\kappa_{\tau}-1)\sum_{f=1}^{V}\sum_{k\in\mathcal{P}}\log{(\tau^{2}_{fk})^{-1}} \\
 - |\mathcal{P}|V\kappa_{\tau}\log\lambda_{\tau} - |\mathcal{P}|V\log\Gamma(\kappa_{\tau}) - \frac{1}{\lambda_{\tau}}\sum_{f=1}^{V}\sum_{k\in\mathcal{P}}(\tau^{2}_{fk})^{-1},
\end{multline*}
where $\mathcal{P}(\mathcal{T})$ is the set of parent topics on the tree. Each draw of $(\kappa_{\tau},\lambda_{\tau})$ is then transformed back to the $(\nu,\sigma^{2})$ scale.

The document-specific topic affinity parameters $\bm{\xi}_{d}$ follow a Multivariate Normal distribution with mean parameter $\bm{\eta}$ and a covariance matrix parameterized in terms of a scalar, $\bm{\Sigma}=\lambda^2\bm{I}_{K}$.
The conditional distribution of these hyperparameters is available in closed form. For efficiency, we choose to put a flat prior on $\log\lambda^{2}$ rather than the original scale, which allows us to marginalize out $\bm{\eta}$ from the conditional posterior of $\lambda^2$:
\begin{eqnarray*}
&\lambda^2 | \{\bm{\xi}_{d}\}_{d=1}^{D} \sim \textrm{Inv-}\chi^{2}\bigg(DK-1,\;\frac{\sum_{d}\sum_{k}(\xi_{dk}-\bar{\xi}_{k})^{2}}{DK-1}\bigg),\\
&\text{and  }\;\;\bm{\eta}|\lambda^2,\{\bm{\xi}_{d}\}_{d=1}^{D} \sim \mathcal{N}\bigg(\bar{\bm{\xi}},\;\frac{\lambda^2}{D}\bm{I}_{K}\bigg).
\end{eqnarray*}

\subsection{Estimation}
As discussed in Section \ref{sec:estimands}, our estimands are the topic-specific frequency and exclusivity of the words in the vocabulary, as well as the FREX score that averages each word's performance in these dimensions. We use posterior means to estimate frequency and exclusivity, computing these quantities at every iteration of the Gibbs sampler and averaging the draws after the burn-in period. For the FREX score, we applied the ECDF function to the frequency and exclusivity posterior expectations of all words in the vocabulary to estimate the true ECDF.

\subsection{Inference for unlabeled documents}
 
In order to classify unlabeled documents, we need to find the posterior predictive distribution of the membership vector $\bm{I}_{\tilde{d}}$ for a new document $\tilde{d}$. Inference is based on the new document's word counts $\bm{w}_{\tilde{d}}$ and the unknown parameters, which we hold constant at their posterior expectation. Unfortunately, the posterior predictive distribution of the topic affinities $\bm{\xi}_{\tilde{d}}$ is intractable without conditioning on the label vector since the labels control which topics contribute content. We therefore use a simpler model where the topic proportions depend only on the relative size of the affinity parameters:
\begin{equation*}
\theta^{*}_{dk}(\bm{\xi}_{d})\equiv\frac{e^{\xi_{dk}}}{\sum_{j=1}^{K}e^{\xi_{dj}}} \;\;\;\text{and }\;\; I_{dk}\sim\text{Bern}\left(\frac{1}{1 + \exp(-\xi_{dk})}\right).
\end{equation*}
The posterior predictive distribution of this simpler model factors into tractable components:
\begin{multline*}
p^{*}(\bm{I}_{\tilde{d}},\bm{\xi}_{\tilde{d}}|\bm{w}_{\tilde{d}},\bm{W},\bm{I}) \approx p(\bm{I}_{\tilde{d}}|\bm{\xi}_{\tilde{d}}) \; p^{*}(\bm{\xi}_{\tilde{d}}|\{\hat{\bm{\mu}}_{f}\}_{f=1}^{V},\hat{\bm{\eta}},\hat{\bm{\Sigma}},\bm{w}_{\tilde{d}}) \\
\varpropto p(\bm{I}_{\tilde{d}}|\bm{\xi}_{\tilde{d}}) \; p^{*}(\bm{w}_{\tilde{d}}|\bm{\xi}_{\tilde{d}},\{\hat{\bm{\mu}}_{f}\}_{f=1}^{V}) \; p(\bm{\xi}_{\tilde{d}}|\hat{\bm{\eta}},\hat{\bm{\Sigma}}).
\end{multline*}
It is then possible to find the most likely $\bm{\xi}^{*}_{\tilde{d}}$ based on the evidence from $\bm{w}_{\tilde{d}}$ alone.

\section{Results}\label{sec-res}
We analyze the fit of the HPC model to Reuters Corpus Volume I (RCV1), a large collection of newswire stories. First, we demonstrate how the variance parameters $\tau^{2}_{fp}$ regularize the exclusivity with which words are expressed within topics. Second, we show that regularization of exclusivity has the greatest effect on infrequent words. Third, we explore the joint posterior of the topic-specific frequency and exclusivity of words as a summary of topical content, giving special attention to the upper right corner of the plot where words score highly in both dimensions. We compare words that score highly on the FREX metric to top words scored by frequency alone, the current practice in topic modeling. Finally, we compare the classification performance of HPC to baseline models.


\subsection{The Reuters Corpus dataset}
RCV1 is an archive of 806,791 newswire stories from a twelve-month period in 1996-1997.\footnote{Available upon request from the National Institute of Standards and Technology (NIST), \texttt{http://trec.nist.gov/data/reuters/reuters.html}} As described in \citet{lewisetal04}, Reuters staffers assigned stories into any subset of 102 hierarchical topic  categories. In the original data, assignment to any topic required automatic assignment to all ancestor nodes, but we removed these redundant ancestor labels since they do not allow our model to distinguish intentional assignments to high level categories from assignment to their offspring. In our modified annotations, the only documents we see in high level topics are those labeled to them and none of their children, which maps onto general content. We preprocessed document tokens with the Porter stemming algorithm (getting 300,166 unique stems) and chose the most frequent $3\%$ of stems (10,421 unique stems, over 100 million total tokens) for the feature set.\footnote{Including rarer features did not meaningfully change the results.}

\begin{figure}[t!]
\caption{Topic hierarchy of Reuters corpus}
\label{reuters-tree}
\begin{center}
\vspace{0.1in}
\includegraphics[width=0.9\textwidth]{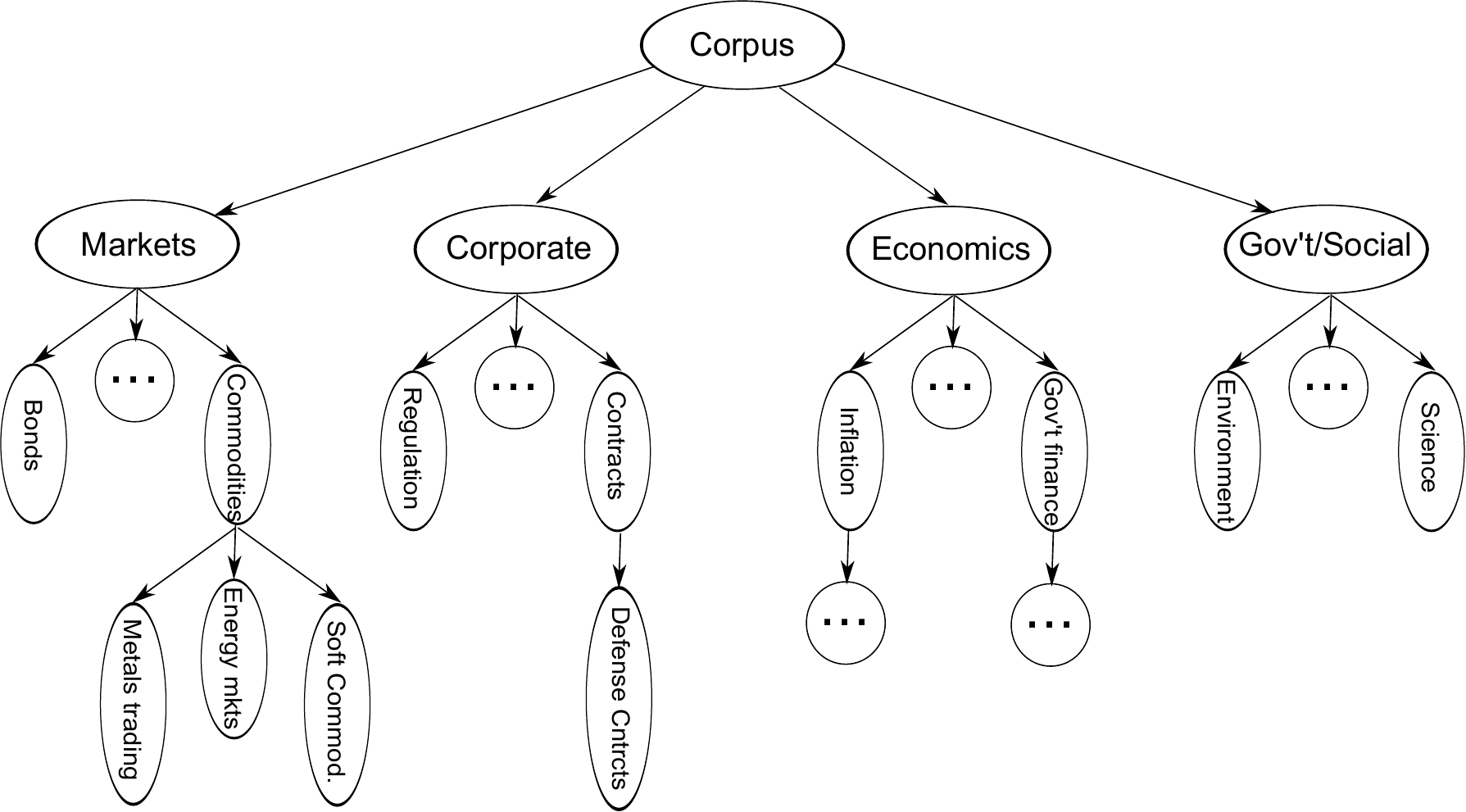}
\end{center}
\end{figure}

The Reuters topic hierarchy has three levels that divide the content into finer categories at each cut. At the first level, content is divided between four high level categories: three that focus on business and market news (Markets, Corporate/Industrial, and Economics) and one grab bag category that collects all remaining topics from politics to entertainment (Government/Social). The second level provides fine-grained divisions of these broad categories and contains the terminal nodes for most branches of the tree. For example, the Markets topic is split between equity, bond, money, and commodity markets at the second level. The third level offers further subcategories where needed for a small set of second level topics. For example, the Commodity Markets topic is divided between agricultural (soft), metal, and energy commodities. We present a graphical illustration of the Reuters topic hierarchy in Figure \ref{reuters-tree}.


Many documents in the Reuters corpus are labeled to multiple topics, even after redundant ancestor memberships are removed. Overall, $32\%$ of the documents are labeled to more than one node of the topic hierarchy. Fifteen percent of documents have very diverse content, being labeled to two or more of the main branches of the tree (Markets, Commerce, Economics, and Government/Social). Twenty-one percent of documents are labeled to multiple second-level categories on the same branch (for example, bond markets and equity markets in the Markets branch). Finally, $14\%$ of documents are labeled to multiple children of the same second-level topic (for example, metals trading and energy markets in the commodity markets branch of Markets). Therefore, a completely general mixed membership model such as HPC is necessary to capture the labeling patterns of the corpus. 
A full breakdown of membership statistics by topic is presented in Tables \ref{docstats1} and \ref{docstats2}.


\begin{table}[ht]
\caption{Topic membership statistics}\label{docstats1}
\scriptsize
\begin{center}
\begin{tabular}{llr....}
  \hline
Topic code & Topic name & \# docs & \multicolumn{1}{c}{\text{Any MM}} & \multicolumn{1}{c}{CB L1 MM} & \multicolumn{1}{c}{CB L2 MM} & \multicolumn{1}{c}{CB L3 MM} \\ 
  \hline
  CCAT & CORPORATE/INDUSTRIAL & 2170 & 79.60\% & 79.60\% & 13.10\% & 0.80\% \\
  C11 & STRATEGY/PLANS & 24325 & 51.50 & 11.50 & 44.50 & 4.50 \\ 
  C12 & LEGAL/JUDICIAL & 11944 & 99.20 & 98.90 & 50.20 & 1.70 \\ 
  C13 & REGULATION/POLICY & 37410 & 85.90 & 55.60 & 61.40 & 4.50 \\ 
  C14 & SHARE LISTINGS & 7410 & 30.30 & 7.90 & 10.30 & 15.80 \\ 
  C15 & PERFORMANCE & 229 & 82.10 & 35.80 & 74.20 & 1.70 \\ 
  C151 & ACCOUNTS/EARNINGS & 81891 & 7.90 & 1.30 & 0.60 & 6.40 \\ 
  C152 & COMMENT/FORECASTS & 73092 & 18.90 & 4.80 & 1.60 & 13.50 \\ 
  C16 & INSOLVENCY/LIQUIDITY & 1920 & 66.70 & 31.50 & 54.60 & 3.60 \\ 
  C17 & FUNDING/CAPITAL & 4767 & 78.10 & 41.40 & 67.70 & 5.00 \\ 
  C171 & SHARE CAPITAL & 18313 & 44.60 & 3.20 & 1.70 & 41.50 \\ 
  C172 & BONDS/DEBT ISSUES & 11487 & 15.10 & 5.70 & 0.30 & 9.70 \\ 
  C173 & LOANS/CREDITS & 2636 & 24.70 & 8.50 & 3.60 & 15.60 \\ 
  C174 & CREDIT RATINGS & 5871 & 65.60 & 59.00 & 0.50 & 7.50 \\ 
  C18 & OWNERSHIP CHANGES &  30 & 76.70 & 23.30 & 76.70 & 3.30 \\ 
  C181 & MERGERS/ACQUISITIONS & 43374 & 34.40 & 6.50 & 4.80 & 26.90 \\ 
  C182 & ASSET TRANSFERS & 4671 & 28.30 & 4.70 & 5.70 & 21.00 \\ 
  C183 & PRIVATISATIONS & 7406 & 73.70 & 34.20 & 6.30 & 44.10 \\ 
  C21 & PRODUCTION/SERVICES & 25403 & 76.40 & 46.50 & 53.60 & 0.80 \\ 
  C22 & NEW PRODUCTS/SERVICES & 6119 & 55.00 & 15.30 & 49.10 & 0.40 \\ 
  C23 & RESEARCH/DEVELOPMENT & 2625 & 77.00 & 36.40 & 57.80 & 0.90 \\ 
  C24 & CAPACITY/FACILITIES & 32153 & 72.20 & 33.60 & 58.40 & 0.90 \\ 
  C31 & MARKETS/MARKETING & 29073 & 46.90 & 25.30 & 34.60 & 1.30 \\ 
  C311 & DOMESTIC MARKETS & 4299 & 80.60 & 73.70 & 9.50 & 18.70 \\ 
  C312 & EXTERNAL MARKETS & 6648 & 78.10 & 70.40 & 9.60 & 14.20 \\ 
  C313 & MARKET SHARE & 1115 & 39.70 & 10.30 & 5.10 & 27.80 \\ 
  C32 & ADVERTISING/PROMOTION & 2084 & 63.80 & 26.90 & 52.50 & 1.40 \\ 
  C33 & CONTRACTS/ORDERS & 14122 & 48.00 & 12.60 & 40.50 & 0.80 \\ 
  C331 & DEFENCE CONTRACTS & 1210 & 68.00 & 65.50 & 13.30 & 3.40 \\ 
  C34 & MONOPOLIES/COMPETITION & 4835 & 92.30 & 54.90 & 75.70 & 14.00 \\ 
  C41 & MANAGEMENT & 1083 & 75.60 & 52.10 & 59.90 & 2.00 \\ 
  C411 & MANAGEMENT MOVES & 10272 & 17.70 & 9.60 & 2.40 & 8.20 \\ 
  C42 & LABOUR & 11878 & 99.70 & 99.60 & 46.50 & 1.50 \\  
  ECAT & ECONOMICS & 621 & 90.50 & 90.50 & 9.70 & 1.40 \\ 
  E11 & ECONOMIC PERFORMANCE & 8568 & 43.00 & 24.20 & 29.10 & 5.10 \\ 
  E12 & MONETARY/ECONOMIC & 24918 & 81.70 & 75.40 & 17.90 & 13.70 \\ 
  E121 & MONEY SUPPLY & 2182 & 30.50 & 23.10 & 0.70 & 9.20 \\ 
  E13 & INFLATION/PRICES & 130 & 60.00 & 46.90 & 28.50 & 0.80 \\ 
  E131 & CONSUMER PRICES & 5659 & 24.70 & 15.60 & 6.00 & 12.00 \\ 
  E132 & WHOLESALE PRICES & 939 & 19.00 & 3.40 & 0.60 & 16.90 \\ 
  E14 & CONSUMER FINANCE & 428 & 73.80 & 43.20 & 61.00 & 1.60 \\ 
  E141 & PERSONAL INCOME & 376 & 75.00 & 63.80 & 9.60 & 22.30 \\ 
  E142 & CONSUMER CREDIT & 200 & 46.00 & 30.00 & 3.50 & 18.50 \\ 
  E143 & RETAIL SALES & 1206 & 27.50 & 19.70 & 2.40 & 10.20 \\ 
  E21 & GOVERNMENT FINANCE & 941 & 86.70 & 81.40 & 53.90 & 4.00 \\ 
  E211 & EXPENDITURE/REVENUE & 15768 & 78.20 & 72.40 & 16.10 & 13.80 \\ 
  E212 & GOVERNMENT BORROWING & 27405 & 32.70 & 29.60 & 2.70 & 4.50 \\ 
  E31 & OUTPUT/CAPACITY & 591 & 45.20 & 18.30 & 35.20 & 0.50 \\ 
  E311 & INDUSTRIAL PRODUCTION & 1701 & 17.70 & 9.80 & 3.10 & 9.30 \\ 
  E312 & CAPACITY UTILIZATION &  52 & 65.40 & 13.50 & 3.80 & 57.70 \\ 
  E313 & INVENTORIES & 111 & 26.10 & 10.80 & 0.00 & 16.20 \\ 
  E41 & EMPLOYMENT/LABOUR & 14899 & 100.00 & 100.00 & 49.40 & 2.20 \\ 
  E411 & UNEMPLOYMENT & 2136 & 92.00 & 90.60 & 10.40 & 12.00 \\ 
  E51 & TRADE/RESERVES & 4015 & 85.10 & 75.50 & 38.70 & 1.90 \\ 
  E511 & BALANCE OF PAYMENTS & 2933 & 63.80 & 43.70 & 8.20 & 25.70 \\ 
  E512 & MERCHANDISE TRADE & 12634 & 64.90 & 59.10 & 11.50 & 11.70 \\ 
  E513 & RESERVES & 2290 & 30.10 & 22.70 & 1.30 & 16.80 \\ 
  E61 & HOUSING STARTS & 391 & 51.70 & 47.80 & 13.80 & 0.80 \\ 
  E71 & LEADING INDICATORS & 5270 & 2.90 & 0.60 & 2.40 & 0.20 \\ 
   \hline
\end{tabular}
\end{center}
\centering \textbf{Key:} MM = Mixed membership, CB Lx = Cross-branch MM at level x
\end{table}
\clearpage

\begin{table}[ht]
\caption{Topic membership statistics, con't}\label{docstats2}
\scriptsize
\begin{center}
\begin{tabular}{llrrrrr}
  \hline
Topic code & Topic name & \# docs & \multicolumn{1}{c}{\text{Any MM}} & \multicolumn{1}{c}{CB L1 MM} & \multicolumn{1}{c}{CB L2 MM} & \multicolumn{1}{c}{CB L3 MM} \\ 
  \hline
  GCAT & GOVERNMENT/SOCIAL & 24546 & 2.50 & 2.50 & 0.50 & 0.10 \\ 
  G15 & EUROPEAN COMMUNITY & 1545 & 16.10 & 6.90 & 14.60 & 0.00 \\ 
  G151 & EC INTERNAL MARKET & 3307 & 98.00 & 87.20 & 10.60 & 94.30 \\ 
  G152 & EC CORPORATE POLICY & 2107 & 96.70 & 90.70 & 40.30 & 50.30 \\ 
  G153 & EC AGRICULTURE POLICY & 2360 & 96.10 & 94.20 & 31.40 & 27.70 \\ 
  G154 & EC MONETARY/ECONOMIC & 8404 & 98.20 & 93.00 & 11.50 & 43.90 \\ 
  G155 & EC INSTITUTIONS & 2124 & 70.80 & 42.00 & 24.30 & 54.00 \\ 
  G156 & EC ENVIRONMENT ISSUES & 260 & 75.00 & 57.70 & 28.80 & 50.80 \\ 
  G157 & EC COMPETITION/SUBSIDY & 2036 & 100.00 & 99.80 & 60.20 & 32.50 \\ 
  G158 & EC EXTERNAL RELATIONS & 4300 & 80.70 & 62.80 & 27.00 & 24.80 \\ 
  G159 & EC GENERAL &  40 & 47.50 & 17.50 & 35.00 & 2.50 \\ 
  GCRIM & CRIME, LAW ENFORCEMENT & 32219 & 79.50 & 41.60 & 59.40 & 0.90 \\ 
  GDEF & DEFENCE & 8842 & 93.70 & 17.20 & 84.40 & 0.50 \\ 
  GDIP & INTERNATIONAL RELATIONS & 37739 & 73.70 & 20.50 & 60.70 & 0.90 \\ 
  GDIS & DISASTERS AND ACCIDENTS & 8657 & 75.70 & 40.10 & 52.20 & 0.20 \\ 
  GENT & ARTS, CULTURE, ENTERTAINMENT & 3801 & 68.80 & 29.20 & 49.60 & 0.50 \\ 
  GENV & ENVIRONMENT AND NATURAL WORLD & 6261 & 90.20 & 51.50 & 72.30 & 2.50 \\ 
  GFAS & FASHION & 313 & 76.40 & 45.70 & 41.50 & 1.90 \\ 
  GHEA & HEALTH & 6030 & 81.90 & 56.10 & 65.00 & 1.20 \\ 
  GJOB & LABOUR ISSUES & 17241 & 99.60 & 99.40 & 44.60 & 3.30 \\ 
  GMIL & MILLENNIUM ISSUES &   5 & 100.00 & 100.00 & 40.00 & 0.00 \\ 
  GOBIT & OBITUARIES & 844 & 99.40 & 15.30 & 99.40 & 0.00 \\ 
  GODD & HUMAN INTEREST & 2802 & 60.70 & 9.70 & 55.20 & 0.10 \\ 
  GPOL & DOMESTIC POLITICS & 56878 & 79.60 & 29.70 & 63.00 & 1.80 \\ 
  GPRO & BIOGRAPHIES, PERSONALITIES, PEOPLE & 5498 & 87.50 & 10.00 & 84.70 & 0.10 \\ 
  GREL & RELIGION & 2849 & 86.10 & 6.60 & 84.30 & 0.10 \\ 
  GSCI & SCIENCE AND TECHNOLOGY & 2410 & 55.20 & 22.20 & 45.10 & 0.30 \\ 
  GSPO & SPORTS & 35317 & 1.30 & 0.60 & 0.90 & 0.00 \\ 
  GTOUR & TRAVEL AND TOURISM & 680 & 89.60 & 69.70 & 34.70 & 3.40 \\ 
  GVIO & WAR, CIVIL WAR & 32615 & 67.30 & 10.10 & 64.60 & 0.10 \\ 
  GVOTE & ELECTIONS & 11532 & 100.00 & 13.30 & 100.00 & 1.30 \\ 
  GWEA & WEATHER & 3878 & 73.90 & 46.80 & 46.40 & 0.10 \\ 
  GWELF & WELFARE, SOCIAL SERVICES & 1869 & 95.40 & 75.50 & 74.10 & 3.40 \\ 
  MCAT & MARKETS & 894 & 81.10 & 81.10 & 14.50 & 2.20 \\ 
  M11 & EQUITY MARKETS & 48700 & 16.30 & 12.30 & 3.90 & 2.90 \\ 
  M12 & BOND MARKETS & 26036 & 21.30 & 15.60 & 5.20 & 3.50 \\ 
  M13 & MONEY MARKETS & 447 & 65.80 & 51.90 & 23.30 & 1.60 \\ 
  M131 & INTERBANK MARKETS & 28185 & 15.10 & 9.40 & 0.70 & 6.40 \\ 
  M132 & FOREX MARKETS & 26752 & 36.90 & 24.70 & 3.10 & 16.10 \\ 
  M14 & COMMODITY MARKETS & 4732 & 18.00 & 16.70 & 2.30 & 0.10 \\ 
  M141 & SOFT COMMODITIES & 47708 & 24.10 & 22.80 & 5.50 & 2.00 \\ 
  M142 & METALS TRADING & 12136 & 34.70 & 19.30 & 4.10 & 16.10 \\ 
  M143 & ENERGY MARKETS & 21957 & 21.10 & 18.40 & 4.80 & 2.90 \\ 
   \hline
\end{tabular}
\end{center}
\centering \textbf{Key:} MM = Mixed membership, CB Lx = Cross-branch MM at level x
\end{table}
\clearpage

\begin{figure}[b!]
\caption{Exclusivity as a function of differential usage parameters}
\label{diff-exc}
\begin{center}
\includegraphics[width=0.49\textwidth]{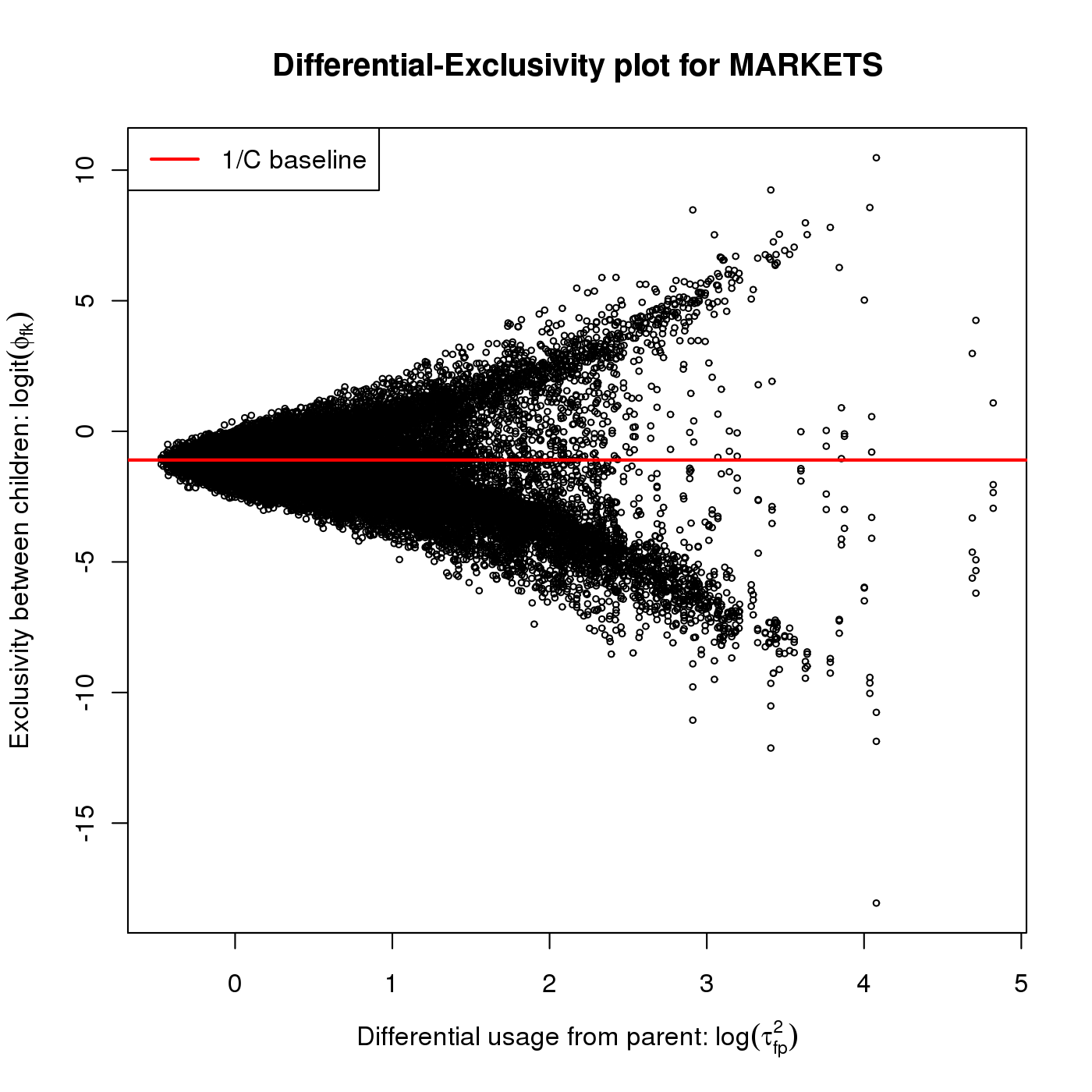}
\includegraphics[width=0.49\textwidth]{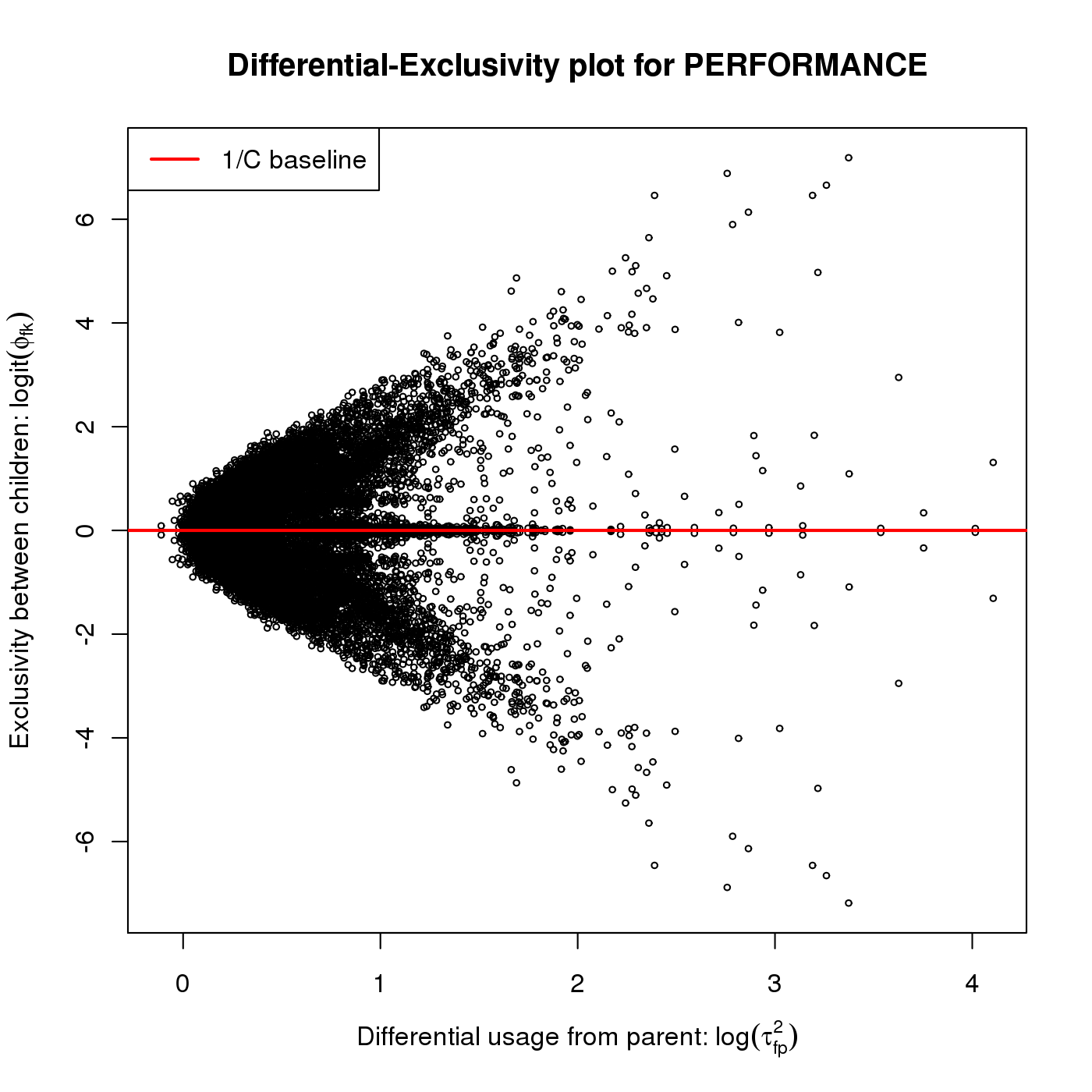}
\end{center}
\end{figure}


\subsection{How the differential usage parameters regulate topic exclusivity}
\label{sec:use-exc}

A word can only be exclusive to a topic if its expression across the sibling topics is allowed to diverge from the parent rate.
Therefore, we would only expect words with high differential usage parameters $\tau^{2}_{fp}$ at the parent level to be candidates for highly exclusive expression $\phi_{fk}$ in any child topic $k$. Words with child topic rates that cannot vary greatly from the parent should have nearly equal expression in each child $k$, meaning $\phi_{fk}\approx\frac{1}{C}$ for a branch with $C$ child topics. An important consequence is that, although the $\phi_{fk}$ are not directly modeled in HPC, their distribution is regularized by positing a prior distribution on the $\tau^{2}_{fp}$.

This tight relation can be seen in the HPC fit. Figure \ref{diff-exc} shows the joint posterior expectation of the differential usage parameters in a parent topic and exclusivity parameters across the child topics. Specifically, the left panel compares the rate variance of the children of Markets from their parent to exclusivity between the child topics; the right panel does the same with the two children of Performance, a second-level topic under the Corporate category. The plots have similar patterns. For low levels of differential expression, the exclusivity parameters are clustered around the baseline value, $\frac{1}{C}$. At high levels of child rate variance, words gain the ability to approach exclusive expression in a single topic.

\subsection{How frequency modulates regularization of exclusivity}
One of the most appealing aspects of regularization in generative models is that it acts most strongly on the parameters for which we have the least information. In the case of the exclusivity parameters in HPC we have the most data for frequent words, so for a given topic the words with low rates should be least able to escape regularization of their exclusivity parameters by our shrinkage prior on the parent's $\tau^{2}_{fp}$.


\begin{figure}[b!]
\caption{Frequency-Exclusivity (FREX) plots}
\label{fe-plot}
\begin{center}
\includegraphics[width=0.45\textwidth]{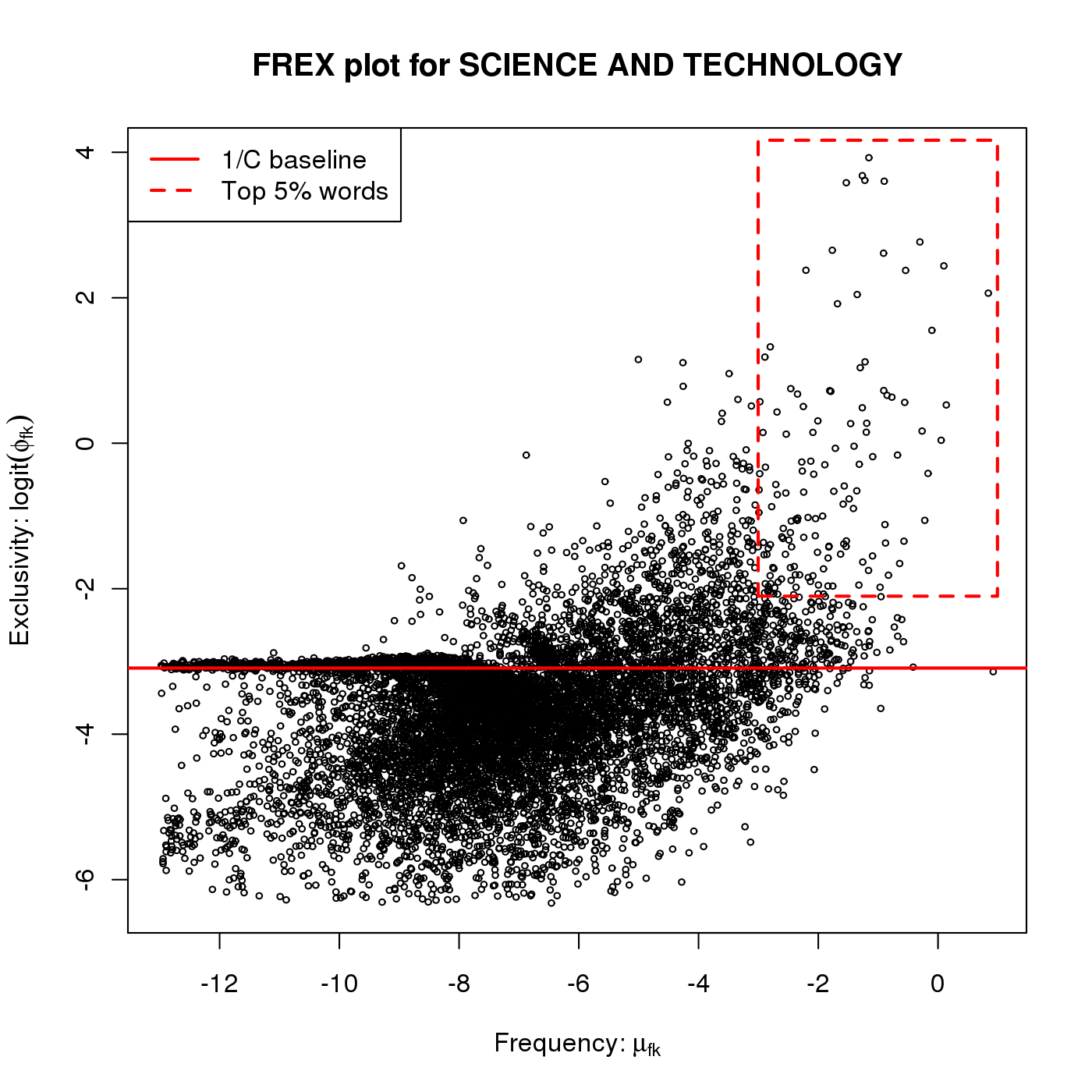}
\includegraphics[width=0.45\textwidth]{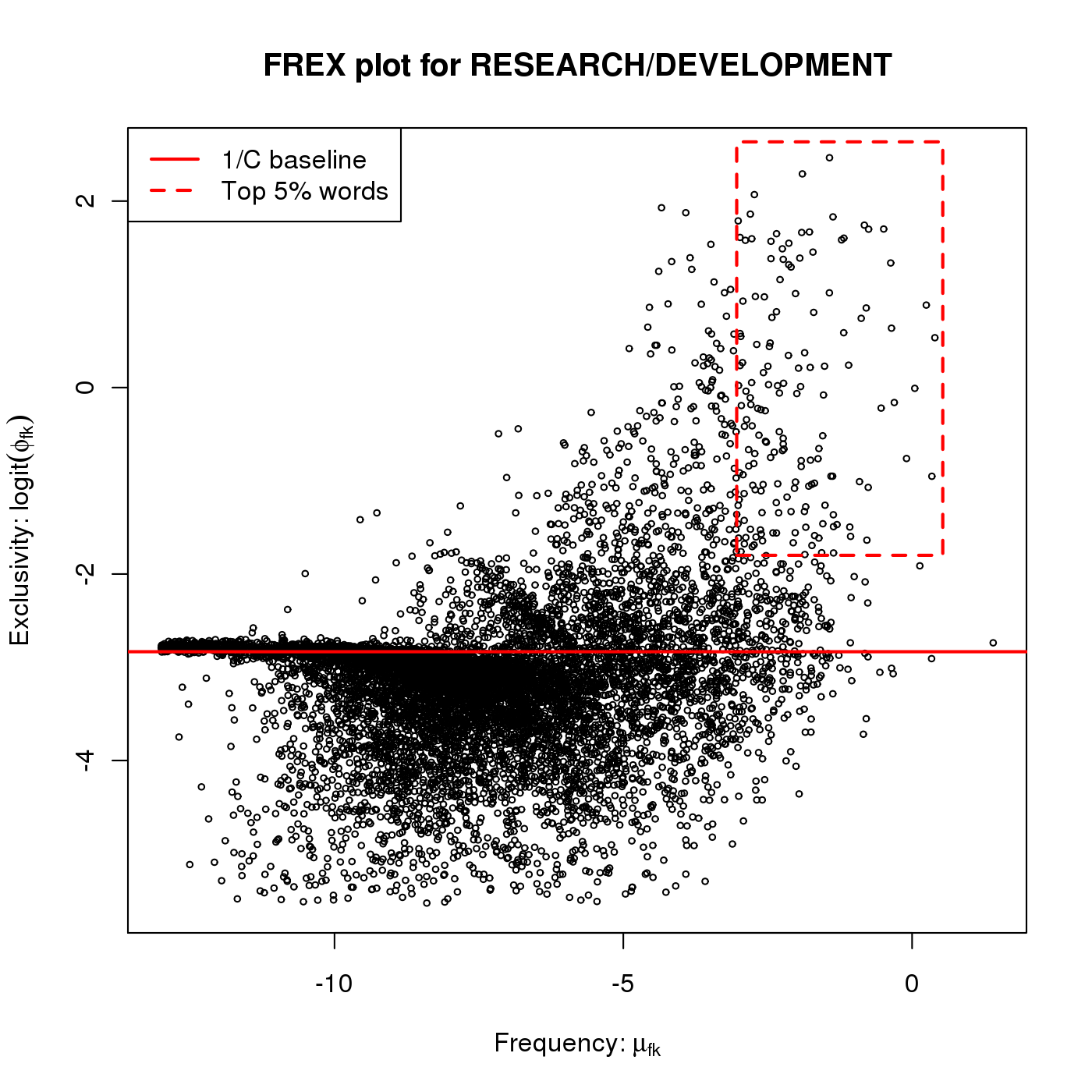}
\end{center}
\end{figure}


\begin{figure}[t!]
\caption{Upper right corner of FREX plot for SCIENCE AND TECHNOLOGY (top) and RESEARCH/DEVELOPMENT (bottom)}
\label{fe-plot-zoom}
\vspace{-0.1in}
\begin{center}
\includegraphics[width=0.67\textwidth]{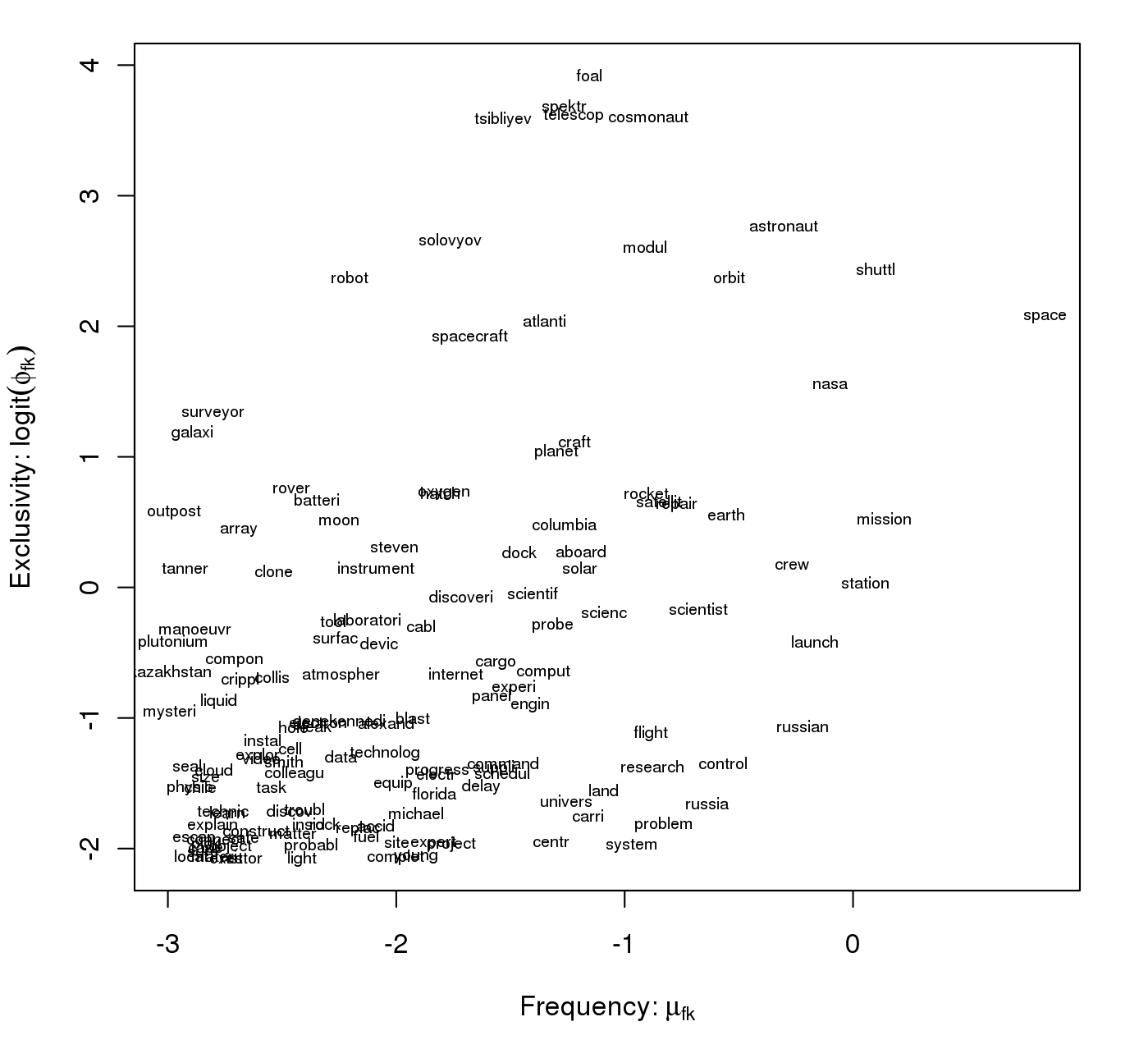}
\includegraphics[width=0.67\textwidth]{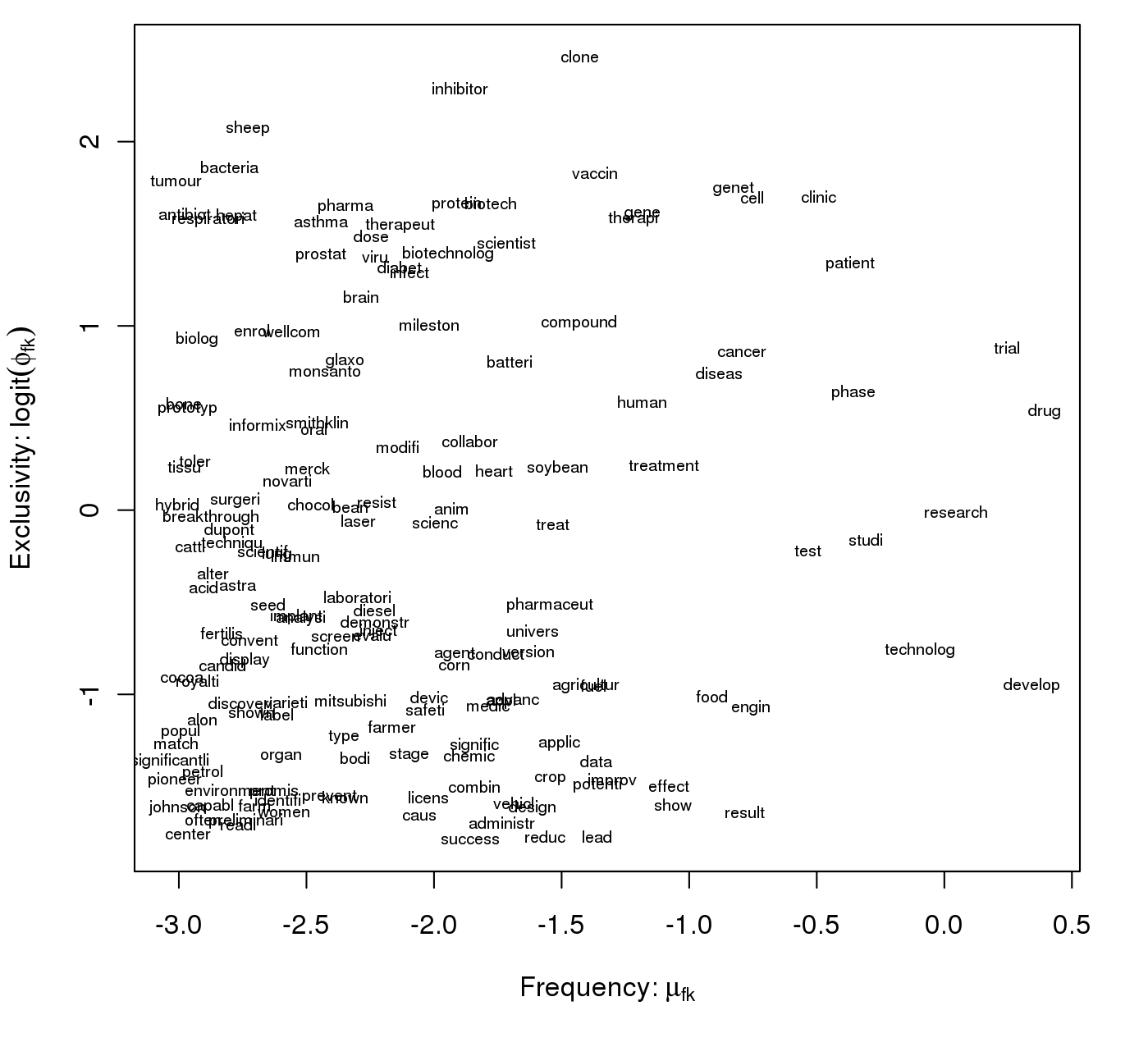}
\end{center}
\end{figure}

Figure \ref{fe-plot} shows for two topics the joint posterior expectation of each word's frequency in that topic and its exclusivity compared to sibling topics (the FREX plot). The left panel features the Science and Technology topic, a child in the grab bag Government/Social branch, and the right panel features the Research/Development topic, a child in the Corporate branch. The overall shape of the joint posterior is very similar for both topics. On the left side of the plots, the exclusivity of rare words is unable to significantly exceed the $\frac{1}{C}$ baseline. This is because the model does not have much evidence to estimate usage in the topic, so the estimated rate is shrunk heavily toward the parent rate. However, we see that it is possible for rare words to be underexpressed in a topic, which happens if they are frequent and overexpressed in a sibling topic. Even though their rates are similar to the parent in this topic, sibling topics may have a much higher rate and account for most appearances of the word in the comparison group.

\subsection{Frequency and Exclusivity as a two dimensional summary of semantic content}
\label{sec:fr-exc}

Words in the upper right of the FREX plot---those that are both frequent and highly exclusive---are of greatest interest. These are the most common words in the corpus that are also likely to have been generated from the topic of interest (rather than similar topics). We show words in the upper $5\%$ quantiles in both dimensions for our example topics in Figure \ref{fe-plot-zoom}. \rev{In particular, words on the left end of these scatterplots are the least frequent, highly exclusive words, and may not appear in  topic summaries  based on frequency alone.} These high-scoring words can help to clarify content even for labeled topics. In the Science and Technology topic, we see almost all terms are specific to the American and Russian space programs. Similarly, in the Research/Technology topic, almost all terms relate to clinical trials in medicine or to agricultural research. 


\begin{table}[ph!]
\small
\caption{Comparison of High FREX words (both frequent and exclusive) to most frequent words (featured topic name bold red; comparison set in solid ovals)}
\label{fe-freq-comp}
\begin{center}
\begin{tabular}{lc|cr}
 & \textbf{High FREX} & \textbf{Most frequent} & \\ 
  \hline
  \multirow{14}{*}{\rotatebox{-90}{\mbox{\textbf{Metals Trading}}}} & copper & said & \multirow{14}{*}{\includegraphics[width=1.25in]{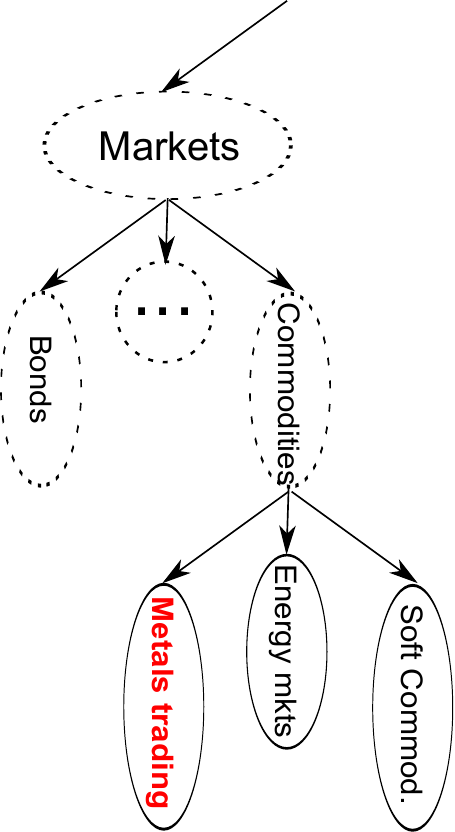}}\\ 
   & aluminium & gold &\\ 
   & metal & price &\\ 
   & gold & copper &\\ 
   & zinc & market &\\ 
   & ounc & metal &\\ 
   & silver & trader &\\ 
   & palladium & tonn &\\ 
   & comex & trade &\\ 
   & platinum & close &\\ 
   & bullion & ounc &\\ 
   & preciou & aluminium &\\ 
   & nickel & london &\\ 
   & mine & dealer &\\ 
  \hline
  \hline
  \multirow{14}{*}{\rotatebox{-90}{\mbox{\textbf{Environment}}}} & greenpeac & said & \multirow{14}{*}{\includegraphics[width=1.25in]{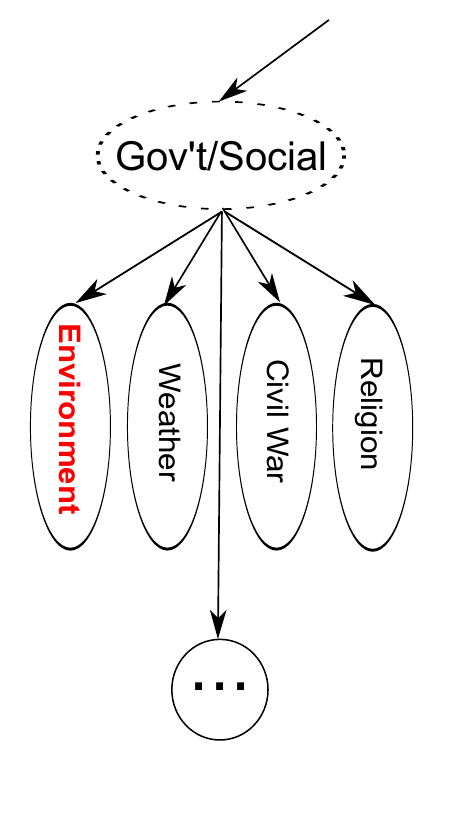}}\\ 
   & environment & would &\\ 
   & pollut & environment &\\ 
   & wast & year &\\ 
   & emiss & state &\\ 
   & reactor & nuclear &\\ 
   & forest & million &\\ 
   & speci & greenpeac &\\ 
   & environ & world &\\ 
   & eleph & water &\\ 
   & spill & group &\\ 
   & wildlif & govern &\\ 
   & energi & nation &\\ 
   & nuclear & environ &\\ 
  \hline
  \hline
  \multirow{14}{*}{\rotatebox{-90}{\mbox{\textbf{Defense Contracts}}}} & fighter & said & \multirow{14}{*}{\includegraphics[width=1.25in]{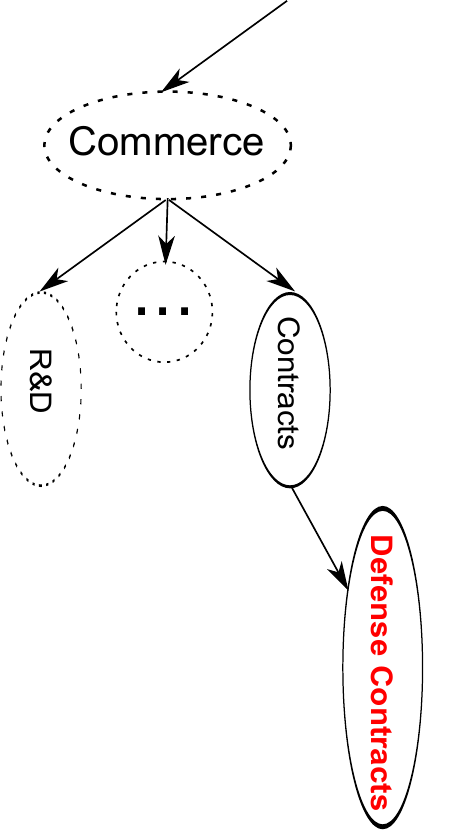}}\\ 
   & defenc & contract &\\ 
   & missil & million &\\ 
   & forc & system &\\ 
   & defens & forc &\\ 
   & eurofight & defenc &\\ 
   & armi & would &\\ 
   & helicopt & aircraft &\\ 
   & lockhe & compani &\\ 
   & czech & deal &\\ 
   & martin & fighter &\\ 
   & militari & govern &\\ 
   & navi & unit &\\ 
   & mcdonnel & lockhe &\\ 
   \hline
\end{tabular}
\end{center}
\end{table}


\begin{figure}[t!]
\caption{Comparison of FREX score components for SMART stop words vs. regular words}
\label{stop-reg-comp}
\begin{center}
\includegraphics[width=0.95\textwidth]{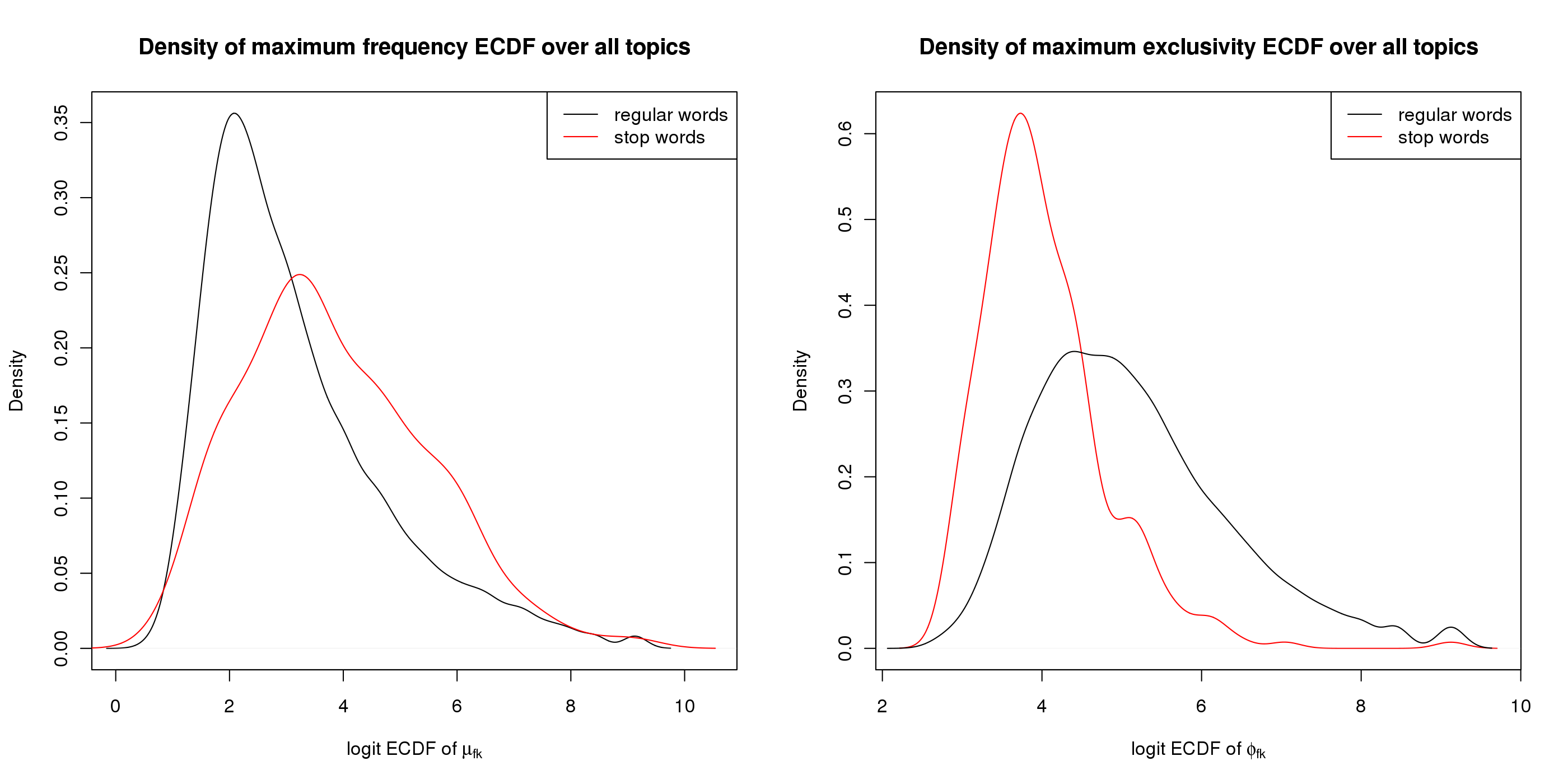}
\end{center}
\end{figure}

We also compute the Frequency-Exclusivity (FREX) score for each word-topic pair, a univariate summary of topical content that averages performance in both dimensions. In Table \ref{fe-freq-comp} we compare the top FREX words in three topics to a ranking based on frequency alone, which is the current practice in topic modeling. For context, we also show the immediate neighbors of each topic in the tree. The topic being examined is in bolded red, while the borders of the comparison set are solid. The Defense Contracts topic is a special case since it is an only child. In these cases, we use a comparison to the parent topic to calculate exclusivity. 

By incorporating exclusivity information, FREX-ranked lists include fewer words that are used similarly everywhere (such as \textit{said} and \textit{would}) and fewer words that are used similarly in a set of related topics (such as \textit{price} and \textit{market} in the Markets branch). One can understand this result by comparing the rankings for known stop words from the SMART list to other words. In Figure \ref{stop-reg-comp}, we show the maximum ECDF ranking for each word across topics in the distribution of frequency (left panel) and exclusivity (right panel) estimates. One can see that while stop words are more likely to be in the extreme quantiles of frequency, very few of them are among the most exclusive words. This prevents general and context-specific stop words from ranking highly in a FREX-based index.  


\subsection{Classification performance}
\rev{We compare the classification performance of HPC with a support vector machine (SVM), a L2-regularized logistic regression and labeled-LDA \citep{ramageetal09}, on both the Reuters corpus and the New York Time corpus \citep{Sandhaus:2008fk}. All methods were trained on a random sample of 15\% of the documents using the $3\%$ most frequent words in the corpus as features. These fits were used to predict memberships in the withheld documents. This out-of-sample prediction experiment was repeated ten times with a new random sample as a training set. 
More in detail, we used a stratified sampling technique to get a balanced sample (across topics) for training, validation, and test partitions with a 15/25/60 split, respectively. We fit the four models to each training set and then used the validation set to calibrate a threshold, except for SVM. We used the fit from the training set and the threshold from the validation set to predict topic memberships in the test set. We trained SVM using both the training and validation data, since it does not need a threshold. 

Table \ref{table-class} shows the results of these experiments, using both micro averages (every document weighted equally) and macro averages (every topic weighted equally).}
\begin{table}[b!]
\rev{
\small
\caption{Classification performance for ten-fold cross-validation. Standard deviation of precision and recall over ten folds is reported in parenthesis.}\label{table-class}
 \begin{center}
  \begin{tabular}{l|llll}
 Data               & \multicolumn{4}{l}{Reuters corpus}                            \\
\belowspace
 Model              & SVM           & L2-reg Logit  & Labeled-LDA   & HPC           \\ \hline
\abovespace
Micro-ave Precision & 0.711 (0.002) & 0.195 (0.031) & 0.487 (0.027) & 0.695 (0.007) \\
\belowspace
Micro-ave Recall    & 0.706 (0.001) & 0.768 (0.013) & 0.543 (0.019) & 0.589 (0.008) \\ \hline
\abovespace
Macro-ave Precision & 0.563 (0.002) & 0.481 (0.025) & 0.344 (0.082) & 0.505 (0.094) \\
\belowspace
Macro-ave Recall    & 0.551 (0.006) & 0.600 (0.007) & 0.476 (0.078) & 0.524 (0.093) \\ \hline
\multicolumn{4}{l}{}\\
 Data               & \multicolumn{4}{l}{New York Times corpus}                            \\
\belowspace
 Model              & SVM           & L2-reg Logit  & Labeled-LDA   & HPC           \\ \hline
\abovespace
Micro-ave Precision & 0.822 (0.001) & 0.378 (0.029) & 0.670 (0.023) & 0.891 (0.003) \\
\belowspace
Micro-ave Recall    & 0.785 (0.001) & 0.732 (0.010) & 0.714 (0.015) & 0.846 (0.004) \\ \hline
\abovespace
Macro-ave Precision & 0.657 (0.002) & 0.512 (0.022) & 0.524 (0.043) & 0.729 (0.054) \\
\belowspace
Macro-ave Recall    & 0.541 (0.004) & 0.580 (0.008) & 0.638 (0.039) & 0.784 (0.043) \\ \hline
 \end{tabular}
 \end{center}}
\end{table}
\rev{HPC compares comparably with SVM on average, dominating on the New York Time corpus, while loosing only to SVM on the Reuters corpus. Labeled-LDA displays a better performance than regularized logistic regression, but loses consistently to both HPC and SVM.
HPC is not designed for optimizing predictive accuracy out-of-sample, rather it is designed to maximize interpretability of the label-specific summaries, in terms of words that are both frequent and exclusive. Neither is labeled LDA. Additional performance gain in prediction tasks for any generative model may be achieved by training such models discriminatively \citep{Zhu:2012fk}.
The results offer a quantitative illustration of the trade-off between predictive and explanatory power of statistical models \citep{Brei:2001b}. For an additional comparative performance evaluation focused on LDA-based models and support vector machines we refer interested readers to \citet{rubinetal12}.}


\rev{
\subsection{Experiments with human evaluators}
\label{sec:mturk}

The data analysis in Sections \ref{sec:use-exc}--\ref{sec:fr-exc} suggest a few hypotheses that warrant further exploration. First, the results suggest FREX summaries improve the interpretability of the topic summaries specified in terms of lists of words. Second, the results suggest the proposed parameterization and regularization scheme for the rates of word occurrence lead to estimates of frequency and exclusivity that are less affected by sampling variations. These in turn translate into further improvements in the interpretability of topic summaries, thus creating a synergistic effect. In addition, in light of these hypotheses, it is plausible to expect that the variability of the estimates of exclusivity are larger, thus less stable, for models that posit regularization of word rates within a topic than for models than in the proposed model, which regularizes the rates of the same word across topics.

In this section we present the results of a large experiment on Amazon Turk that enlists human evaluators to test the hypotheses we outlined above quantitatively.

\subsubsection{Design choices}

In order to avoid confounding issues we consider two models in this experiment: Latent Dirichlet Allocation \citep[][]{bleietal03}, the simplest model with that posit regularization of word rates within a topic, and a simpler unsupervised variant of the proposed model that regularizes the rates of the same word across topics, without the hierarchy on the rates of word occurrence.

We quantify interpretability in two ways: indirectly, in terms of topic diversity, measured by the number of unique words that appear in 5-, 10-, 25- and 5-word topic summaries, and directly, in terms of evaluator preferences for the topic summary of a model over that of another.
We quantified stability of the estimates of exclusivity both directly, in terms of variance of the estimated word rates, and indirectly, in terms of the maximum exclusivity of a word across topics.

We fit  both models to 2,246 documents from Academic Press corpus \citep{harman92}. We performed some pre-processing steps to maximize the interpretability of inferred topics, and to avoid confounding in the results due to the different impact of stop words on the word rates inferred with different models. First, we removed all stop words on the SMART stop list to prevent obvious filler words from dominating topic summaries.\footnote{This list is available at \href{http://www.jmlr.org/papers/volume5/lewis04a/a11-smart-stop-list/english.stop}{www.jmlr.org/papers/volume5/lewis04a/a11-smart-stop-list/english.stop}.} We also removed all proper nouns using the part-of-speech tagger in the Python Natural Language Toolkit (NLTK) so that topic summaries do not require encyclopedic knowledge of people and places to be interpretable. For both models we varied the number of topics across a wide range, with $K\in\{10,25,50,100\}$.

We sought to compare 
 FREX-based topic summaries obtained with the proposed model, 
 to topic summaries obtained with LDA, 
 to FREX-based topic summaries where the estimates of frequency and exclusivity are obtained by leveraging LDA word rate estimates---thus effectively using the proposed FREX score to re-rank the words associated with each topic according to LDA.

The latent Dirichlet allocation model is parameterized in terms of the probability of a word given a topic, not the topic given a word. In order compute the FREX score from LDA word rate estimates, we need to reverse this conditioning. A simple calculation involving the marginal probability of each topic is necessary. Specifically,
\begin{equation}
\label{ustext:eq:lda-norm}
p(\text{topic } k|\text{word } f) = \frac{p(\text{word } f|\text{topic } k)p(\text{topic } k)}{\sum_{j=1}^{K}p(\text{word } f|\text{topic } j)p(\text{topic } j)}.
\end{equation} 
Since LDA uses a symmetric Dirichlet prior on the topic membership probabilities, the marginal topic probabilities are equal. Therefore the conditional distributions are equal and no correction is needed. However, for more complicated models where topic probabilities can be unequal \citep[e.g., see][]{blei:2012}, a posterior estimate of this inverse probability is required to get the FREX score. 

\subsubsection{Diversity in the inferred topics}
\label{sec:diversity}

The analysis in Section \ref{sec:fr-exc} suggested that the FREX score helps produce more diverse topical summaries. A set of topics that do not overlap in their word summaries are arguably provides a more interpretable thematic structure underlying a given collection of documents.
Here, we compare the diversity of topical summaries obtained with the proposed approach and with LDA.

One simple metric for quantifying the diversity between topic summaries is the proportion of unique words across all the summaries produced from a model fit. For example, five-word summaries from a 100-topic model would have at most 500 unique words, and the proportion of the total achieved is an indication for whether the word lists are presenting diverse information. 
Table \ref{ustext:fig:prop-unique} shows this proportion for 5-, 10-, 25-, and 50-word summaries obtained wither strategies of interest: 
 ranking words by FREX scores estimated using the proposed Poisson convolution model (PCM FREX in the table),
 re-ranking words by FREX score estimated leveraging LDA word rate estimates (LDA FREX in the table), and 
 ranking words by frequency using LDA word rate estimates (LDA FREQ in the table).

\begin{table}[t!]
\rev{
\caption{Proportion of unique words in topic summaries obtained with different strategies}
\label{ustext:fig:prop-unique}
\begin{center}
\begin{tabular}{rrrrr}
    \multicolumn{5}{c}{(a) 5-word summaries} \\
  \hline
  N topics & 10 & 25 & 50 & 100 \\ 
  \hline
  PCM FREX & 1.000 & 1.000 & 1.000 & 0.998 \\ 
  LDA FREX & 1.000 & 1.000 & 1.000 & 0.974 \\ 
  LDA FREQ & 0.820 & 0.752 & 0.612 & 0.522 \\ 
   \hline
 \vspace{0.5em} \\
    \multicolumn{5}{c}{(b) 10-word summaries} \\
    \hline
    N topics & 10 & 25 & 50 & 100 \\ 
  \hline
PCM FREX & 1.000 & 1.000 & 0.998 & 0.989 \\ 
  LDA FREX & 1.000 & 1.000 & 0.990 & 0.948 \\ 
  LDA FREQ & 0.790 & 0.744 & 0.594 & 0.462 \\ 
   \hline
   \vspace{0.5em} \\
    \multicolumn{5}{c}{(c) 25-word summaries} \\
  \hline
  N topics & 10 & 25 & 50 & 100 \\ 
  \hline
  PCM FREX & 1.000 & 0.998 & 0.978 & 0.924 \\ 
  LDA FREX & 1.000 & 0.997 & 0.942 & 0.846 \\ 
  LDA FREQ & 0.744 & 0.650 & 0.493 & 0.384 \\ 
   \hline
      \vspace{0.5em} \\
    \multicolumn{5}{c}{(d) 50-word summaries} \\
  \hline
  N topics & 10 & 25 & 50 & 100 \\ 
  \hline
  PCM FREX & 1.000 & 0.997 & 0.977 & 0.907 \\ 
  LDA FREX & 1.000 & 0.985 & 0.934 & 0.826 \\ 
  LDA FREQ & 0.678 & 0.553 & 0.448 & 0.384 \\ 
   \hline
\end{tabular}
\end{center}}
\end{table}

The results show that topic summaries obtained with the proposed model in combination with the FREX score lead to over $90\%$ the words begin unique, independently of the number of words in the summary, and of the number of topics in the range we considered---which is typical for topic models. When restricting to 5- and 10-word summaries almost all the words are unique. 
For frequency-based LDA summaries, in contrast, the proportion of unique words drops off quickly as the length of the summaries and number of topics increases. For example, even in 5-word summaries of a 100-topic model, only half the words are unique to any given topic. This repetition makes it more difficult to understand distinct thematic concepts reflected in each topic and may reduce the interpretability of the model fit. 
Using a FREX-based summary with LDA word rate estimates to re-rank topic summaries does increase the proportion of unique words. The gains in topic diversity, however, become less pronounced both as the length of the summaries and the number of topics increase.
These results can be explained, in part, by the fact that FREX-based summaries do not share non-topical ``filler'' words across topics, which can dominate their frequency-based summaries. FREX scores  also increase the  diversity in the topic summaries by promoting less common words that only occur in a given topic. 

In the next section, we use human evaluators to determine the extent to which topic summaries containing a larger fraction of unique words convey more interpretable themes.

\subsubsection{A randomized experiment to compare the interpretability of topic summaries}

The two compelling hypotheses that the previous experiments and analyses suggest fairly strongly are that topic summaries based on the FREX score are more interpretable than currently established frequency based summaries, and that the proposed model produces estimates of the FREX scores that are superior to those obtained from LDA. However, interpretability hard to quantify, and it is difficult to develop automated methods that are reliable proxies for human judgement. For instance, recent research has found that out-of-sample likelihood is negatively correlated with human judgements of topic interpretability \citep{changetal09}. 
Here we repot the results of large randomized experiment we conducted on Amazon Mechanical Turk (\href{http://aws.amazon.com/mturk/}{aws.amazon.com/mturk}) that aims at leveraging human evaluators to execute a comparative analysis of the interpretability of topic summaries, obtained with the three different strategies we have been considering. The experiment consists of evaluation tasks that require participants to interact with the output of different models, in a way that tests their ability to extract coherent themes from the topic summaries these models produce \citep{Newman2010,Aletras2013,Jia:2014fk}.

We implement two human evaluation tasks with users from Amazon Turk that both involve comparing the threes strategies of interest for producing topic summaries (PCM FREX, LDA FREQ and LDA FREX, using abbreviations established in Section \ref{sec:diversity}) to test the two hypotheses about model interpretability outlined above. 
We refer to the first task as the ``word intrusion" task \citep{changetal09}. This task measures the coherence of topic summaries by asking users to find which word does not belong in a topic summary; the  intruder word is chosen among the words that are highly associated with another topic. Intuitively, intruder words will be easiest to identify in summaries that express clear and distinct themes. 
We refer to the second task as the ``topic coherence" task \citep{Newman2010}. This tasks  involves directly asking users to rate the coherence of a topic summary on a 1-3 scale. In addition, to get a clearer picture of the relative value of the three strategies to produce topic summaries we consider, after asking users to rate summaries from each of the methods, we also ask them to identify the most coherent summary among them, with an option for stating they have no preference. 

\begin{figure}[p]
\rev{\begin{center}
(a) Word intrusion example
\fbox{\includegraphics[width=0.8\textwidth]{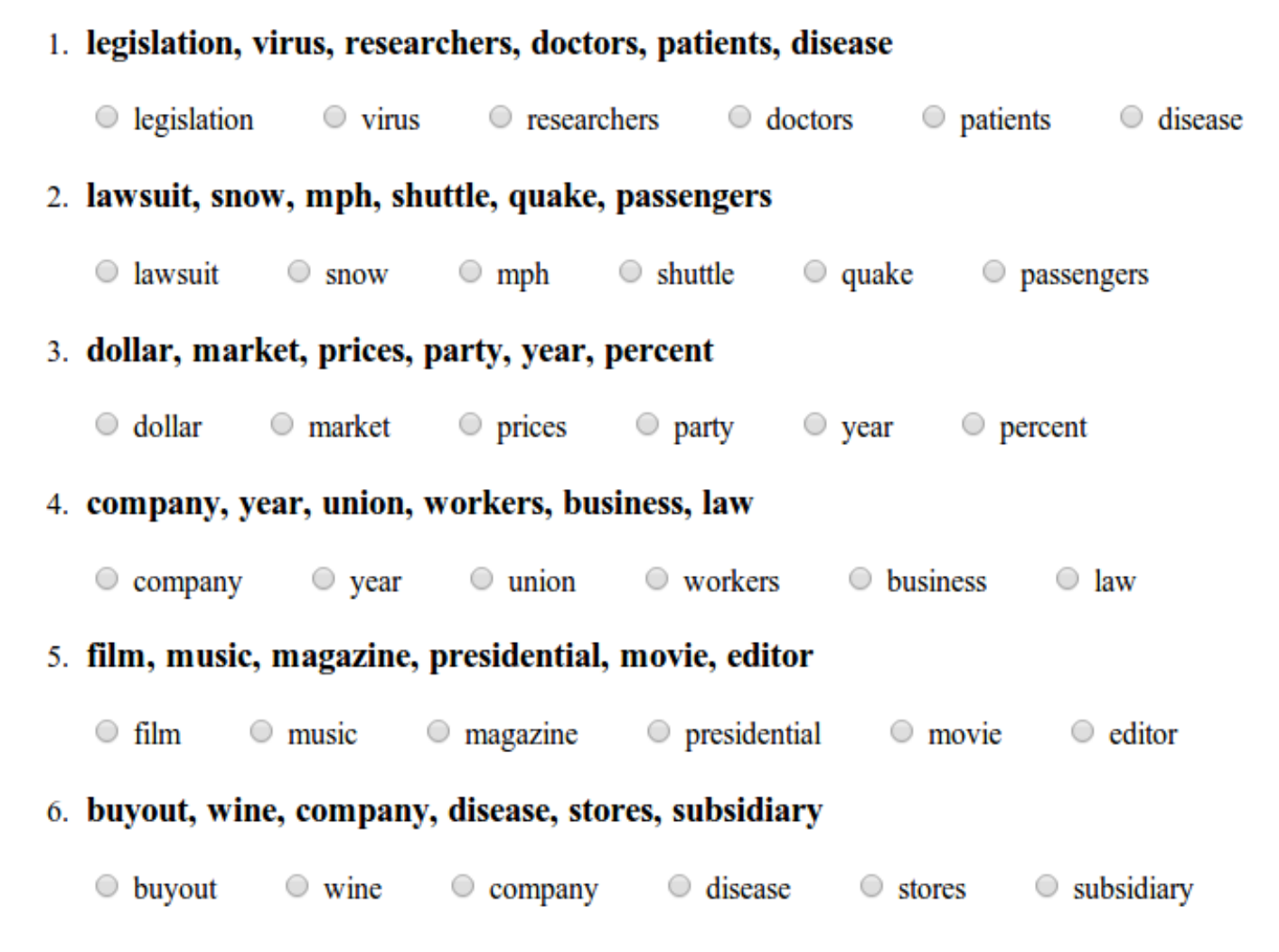}}
\end{center}
\begin{center}
(b) Topic coherence example
\fbox{\includegraphics[width=0.8\textwidth]{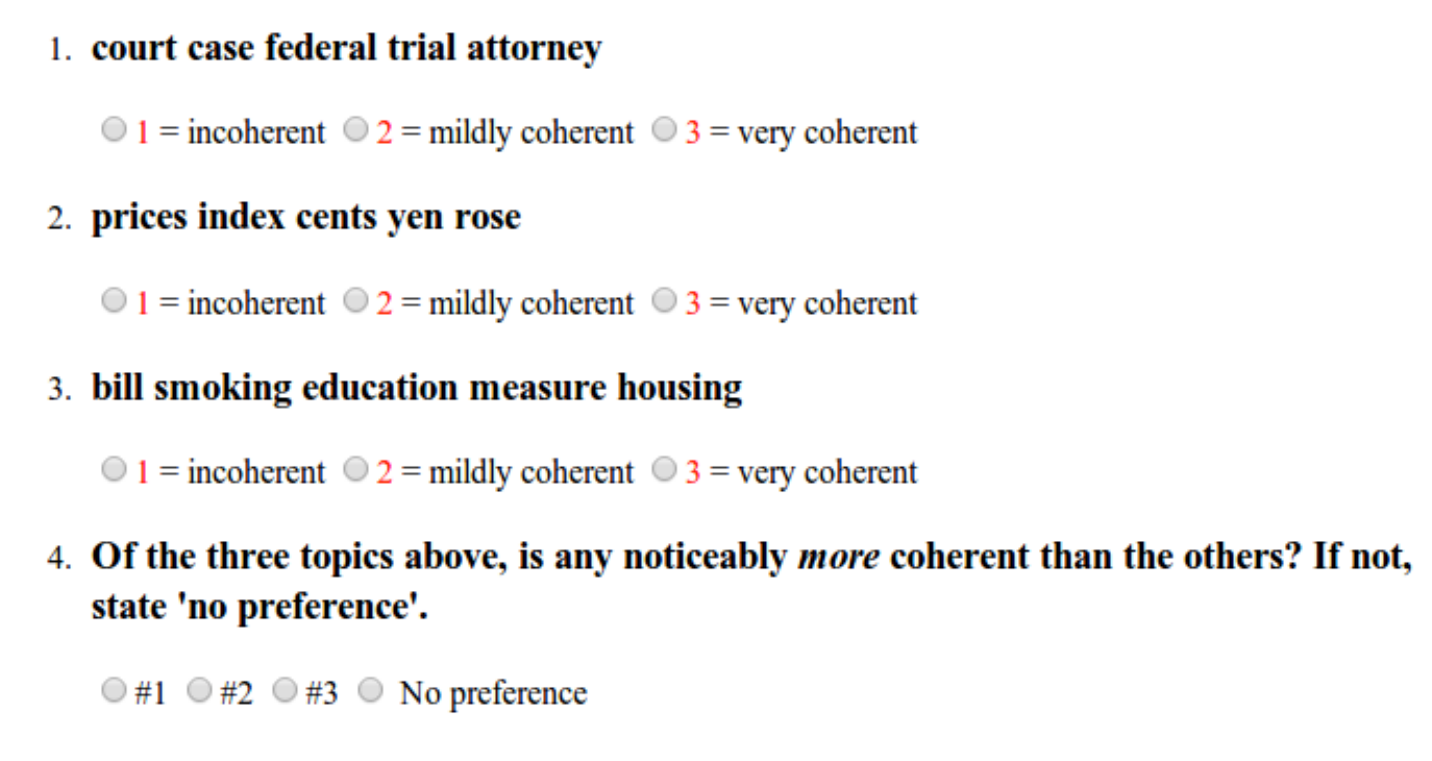}}
\end{center}
\caption{Screenshots of Amazon Turk tasks}}
\label{ustext:fig:aturk_example}
\end{figure}

Figure \ref{ustext:fig:aturk_example}a shows an example of a word intrusion task. 
Each of the questions presents---for a single strategy---the top five scoring words in a random topic and along with an intruder word from the top twenty scoring words in one of the other topics. The order of words in the list is shuffled randomly before being presented to the user, who is asked to identify the intruder. Each task has six questions---exactly two from each strategy---also presented in a random order. All the 5-word topic summaries being compared in a task come from models with the same number of topics. The estimand of interest is the probability of correctly identifying the intruder word associated with each strategy to produce topic summaries. We considered models with 10, 25, 50 and 100 topics. For each model size, we gave the task to 400 users, resulting in a total 3,200 responses for each of the strategy to produce topic summaries.

Figure \ref{ustext:fig:aturk_example}b shows an example of a topic coherence task. 
The first three questions provide a randomly chosen summary from each of the strategies and asks the user to rate it on a 1-3 scale. The order of summaries is randomized. Several examples of coherent and incoherent topics are provided to users in an included rubric. The final question asks the user if any of the three summaries are noticeably more coherent than the others to gauge the relative interpretability of the strategies. Included is an option to express no preference so that users do not choose arbitrarily whenever there is not an obvious top choice. The two estimands of interest are the average rating for each type of strategy, and the probability each strategy being the most coherent. We considered models with 10, 25, 50 and 100 topics. For each model size, we gave the task to 400 users, resulting in a total 1,600 responses about absolute coherence and 1,600 responses about relative coherence  for each of the strategy to produce topic summaries.

\begin{figure}[t!]
\rev{\begin{center}
\includegraphics[width=0.9\textwidth]{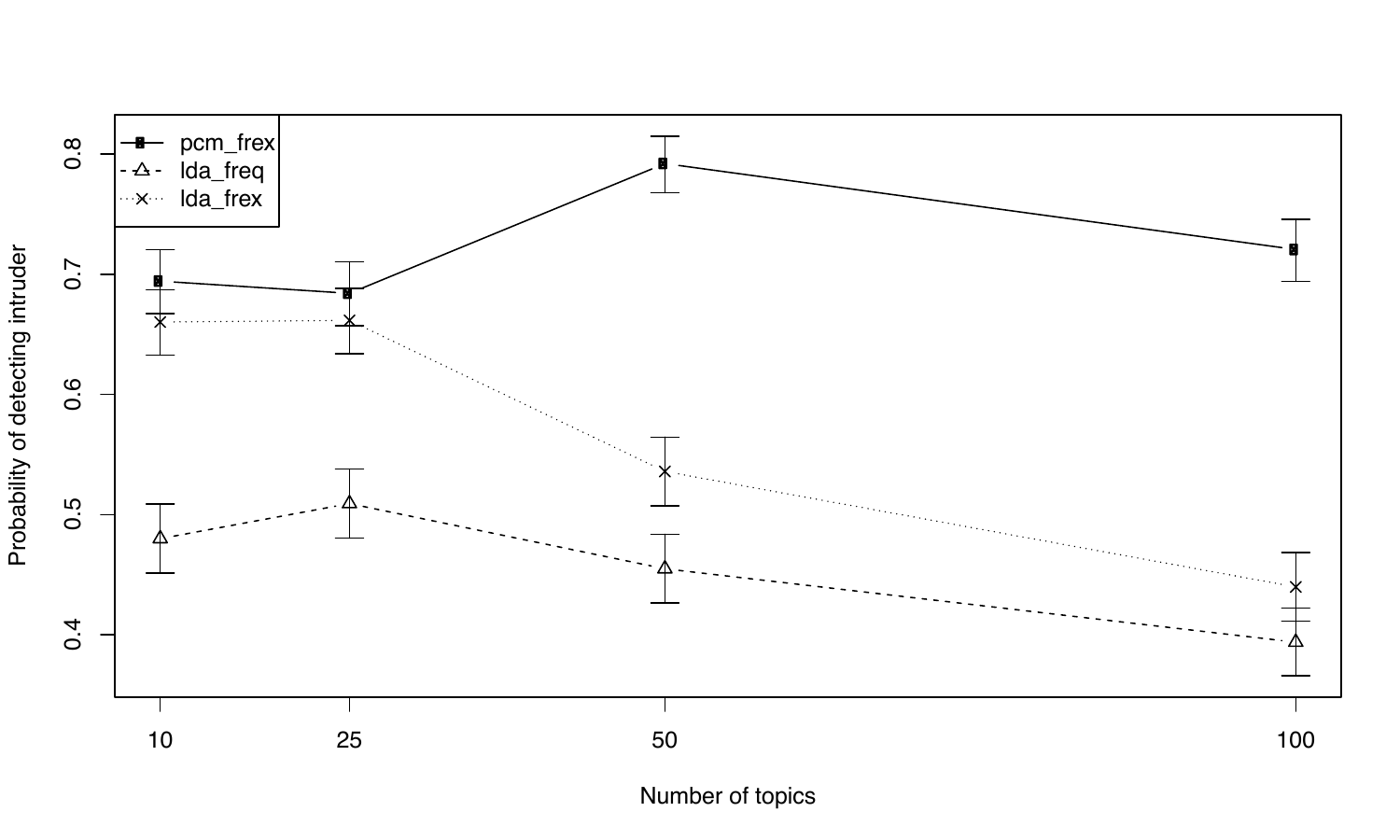}
\end{center}
\caption{Results from Amazon Turk word intrusion task}}
\label{ustext:fig:aturk-res-wi}
\end{figure}

\begin{figure}[t!]
\rev{\begin{center}
(a) Average ratings for individual summaries \\
\vspace{-2em}
\includegraphics[width=0.9\textwidth]{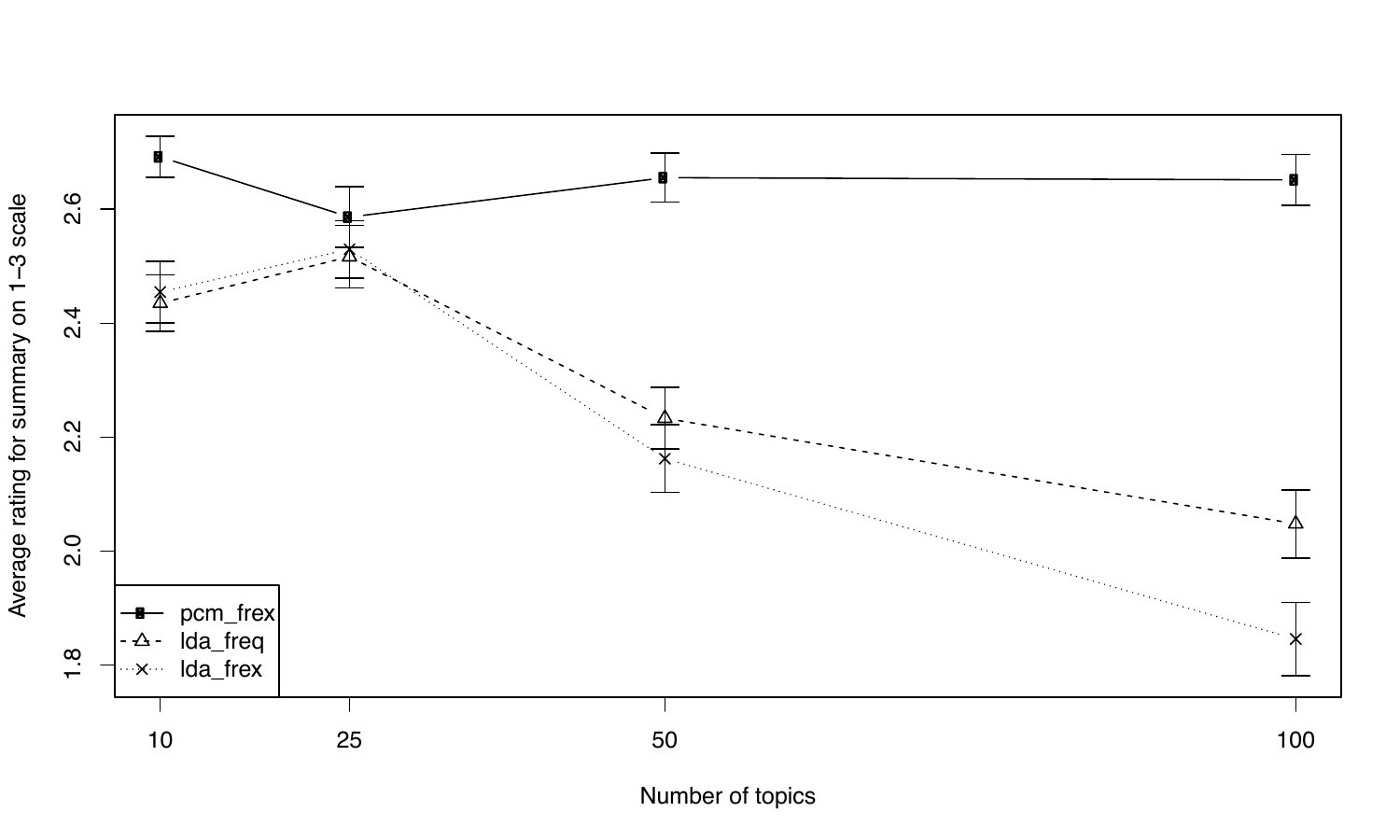} \\
(b) Relative preference across summary methods \\
\vspace{-2em}
\includegraphics[width=0.9\textwidth]{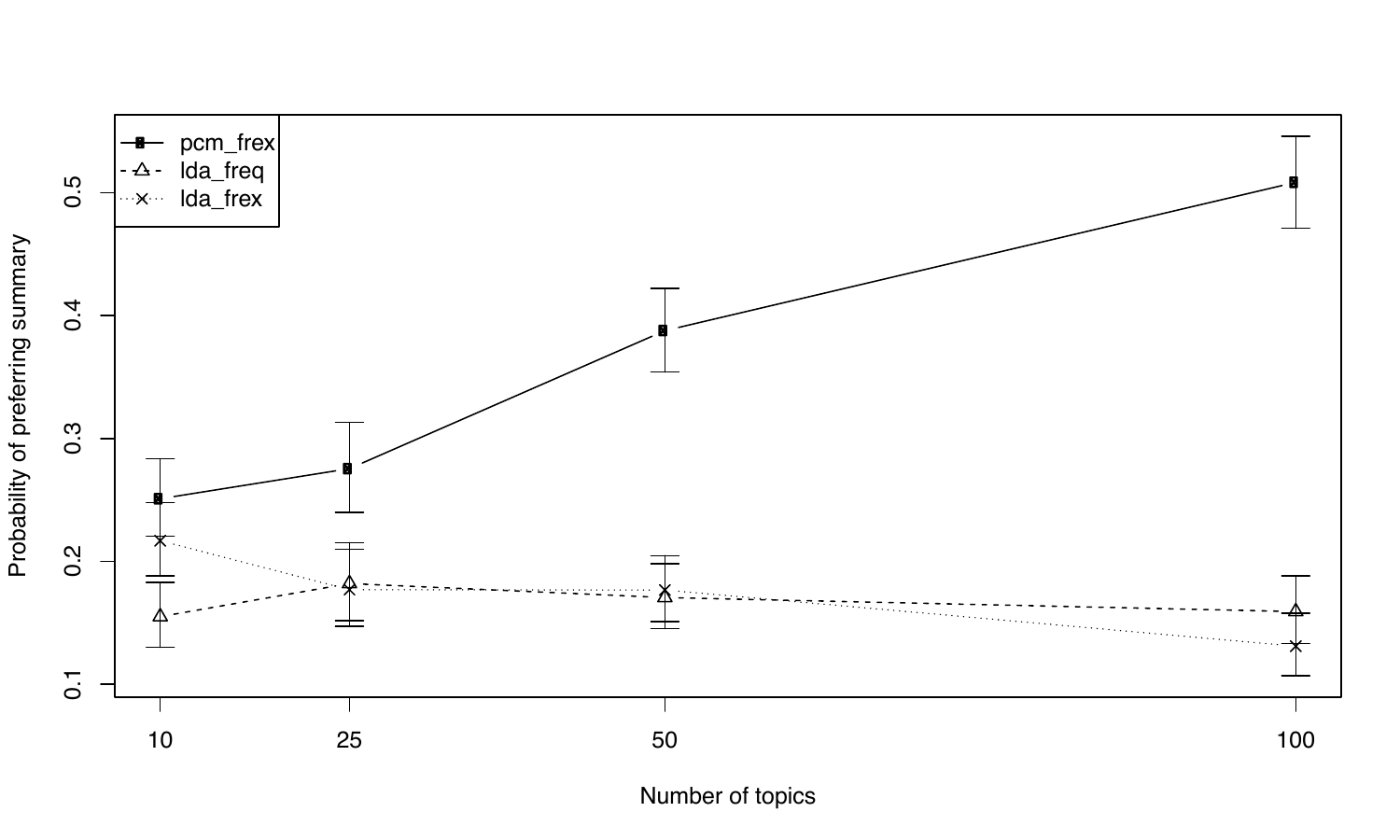}
\end{center}
\caption{Results from Amazon Turk topic coherence task}}
\label{ustext:fig:aturk-res-tc}
\end{figure}

Figure \ref{ustext:fig:aturk-res-wi} shows the results for the word intrusion task. 
In the plot we compare the probability of a user finding the intruder word across all three strategies as a function of the number of topics in the model. The performance for frequency based summaries using LDA is consistently low, with the detection probability at $0.5$ for small topic spaces and falling to $0.4$ for a 100-topic model. Re-ranking LDA topic summaries using the FREX scores only improves performance for models with small number of topic, with the probability of finding the intruder word nearly equal that of the LDA frequency based summary for models with 50 and 100 topics. In contrast, users consistently detect the intruding words with high probability---between 0.7 and 0.8---when the topic summaries are based on the FREX score estimated using the proposed model. Furthermore, these results indicate that the interpretability of these topic summaries does not degrade as the size of the number of topic in the model increases.  

Figure \ref{ustext:fig:aturk-res-tc} shows the results for the topic coherence task. 
Panel (a) shows the average absolute coherence ratings of the topic summaries obtained with each of the three strategies. Similar to the word intrusion results, the summaries produced by the CPM FREX strategy maintain consistently high ratings---around 2.6 out of 3---regardless of the size of the model. Interestingly, topic summaries obtained with both strategies based on LDA, whether ranked according to frequency or to FREX, lead to indistinguishable ratings for most model sizes, with high ratings for smaller models that quickly drop as the size of the models increase. For the model with  100 topics, re-ranking by LDA FREQ topic summaries by FREX scores display the worst performance, with average ratings below two.
Panel (b) of Figure \ref{ustext:fig:aturk-res-tc} shows the relative preferences of human evaluators for the three strategies to produce topic summaries. Again, the preference for the topic summaries produced by the FREX scores estimated with the proposed method (PCM FREX) increases as the size of the models increase, with over $50\%$ of workers choosing that strategy for the largest model we considered, with 100 topics. Interestingly, preference for the topic summaries obtained with both strategies based on LDA, whether ranked according to frequency or to FREX, is consistently low independently of the model size. 

\begin{figure}[b!]
\rev{\begin{center}
\includegraphics[width=0.9\textwidth]{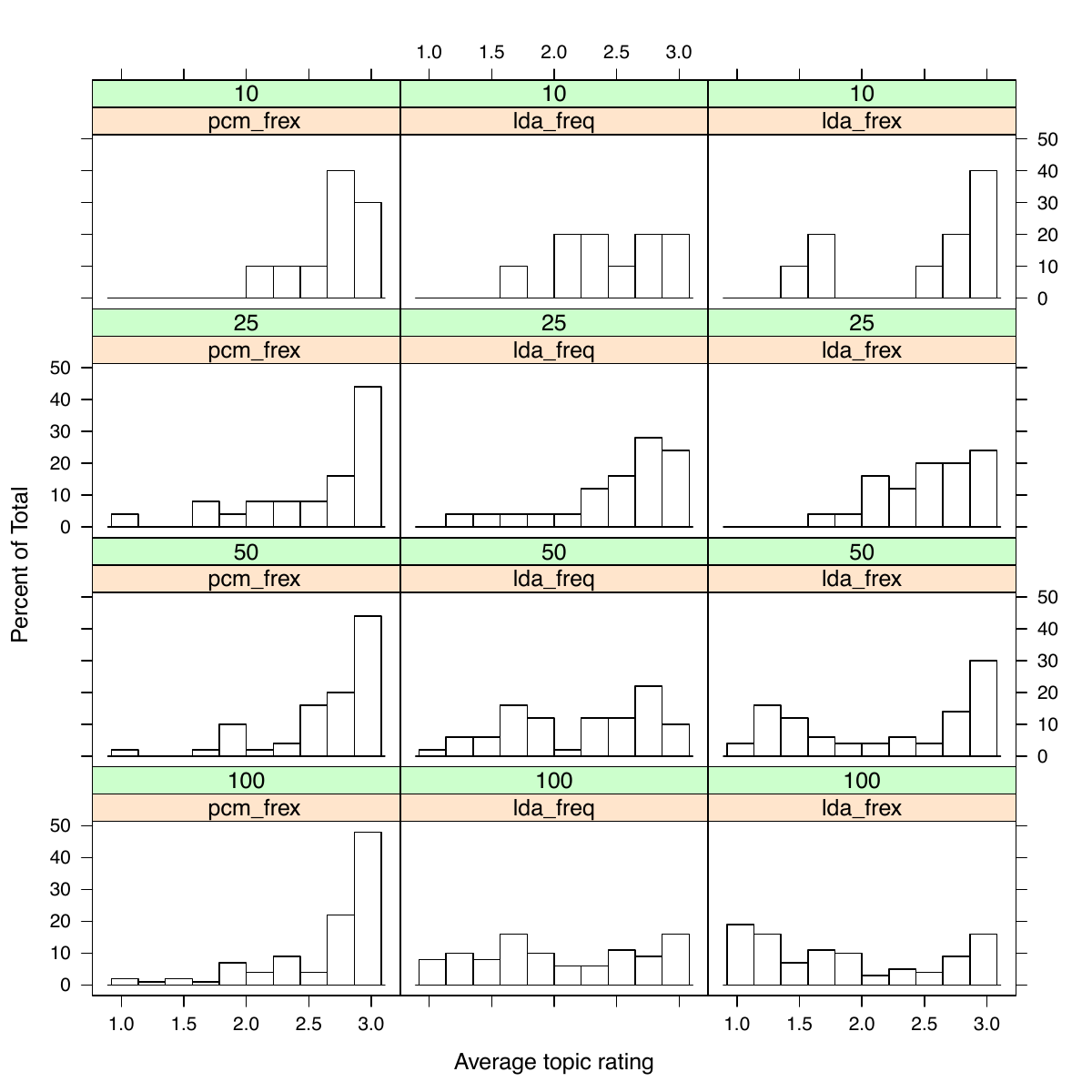}
\end{center}
\caption{Distribution of topic coherence ratings across number of topics in model (rows) and summary method (columns)}}
\label{ustext:fig:aturk-topic-dist}
\end{figure}

A natural question arising from the results in Figures \ref{ustext:fig:aturk-res-wi} and \ref{ustext:fig:aturk-res-tc} is whether the average quality degradation we observe for topic summaries based on LDA as the number of topics in the model grows is due to the declining quality of all topics or  to the addition of many low quality topics. In order to better understand this observed trend in average quality, Figure \ref{ustext:fig:aturk-topic-dist} shows the distribution of average coherence ratings for individual topic summaries obtained with the three strategies. In the Figure,  the number of topics in the model varies along the rows, while the strategy to obtain topic summaries varies along the columns. The distributions of topic coherence ratings from human evaluators for LDA FREQ (middle column) and LDA FREX (right column) flatten out for larger models, rather than concentrating around a mediocre score. This suggest that while some high-quality topic summaries remain for the lager 50- and 100-topic models, these high-quality topic are outnumbered by a growing number of middle- and low-quality topic summaries. In contrast, the distributions of the ratings for PCM FREX topic summaries remains relatively unchanged as the models grow in size, with most of the topic summaries retaining average ratings above 2.5. 


Overall, the results of the randomized experiment on Amazon Mechanical Turk provide strong evidence in support of the two hypotheses that 
 topic summaries based on the FREX score are more interpretable than currently established frequency based summaries, and 
 that the proposed model produces estimates of the FREX scores that are superior to those obtained from LDA. 

\subsubsection{Stability of exclusivity estimates}
\label{sec:stability} 

The experiments above suggest that, while the exclusivity of a word to a topic can be computed from word rate estimates obtained with an LDA type parameterization, such estimates lead to less interpretable topics, especially in larger models. One plausible explanation for these results is that estimates of exclusivity based on models that regularize word rates within a topic are less stable, in some sense, than estimates obtained with the proposed approach to modeling and regularization.
Here, we explore the stability of the exclusivity estimates  obtained with both approaches. 


Stability is quantified
 indirectly, in terms of the maximum exclusivity of a word across topics, and 
 directly, in terms of variance of the estimated word rates.
\begin{figure}[p]
\rev{
\begin{center}
Regularization within topic \hspace{90pt} Regularization across topics
\includegraphics[width=\textwidth]{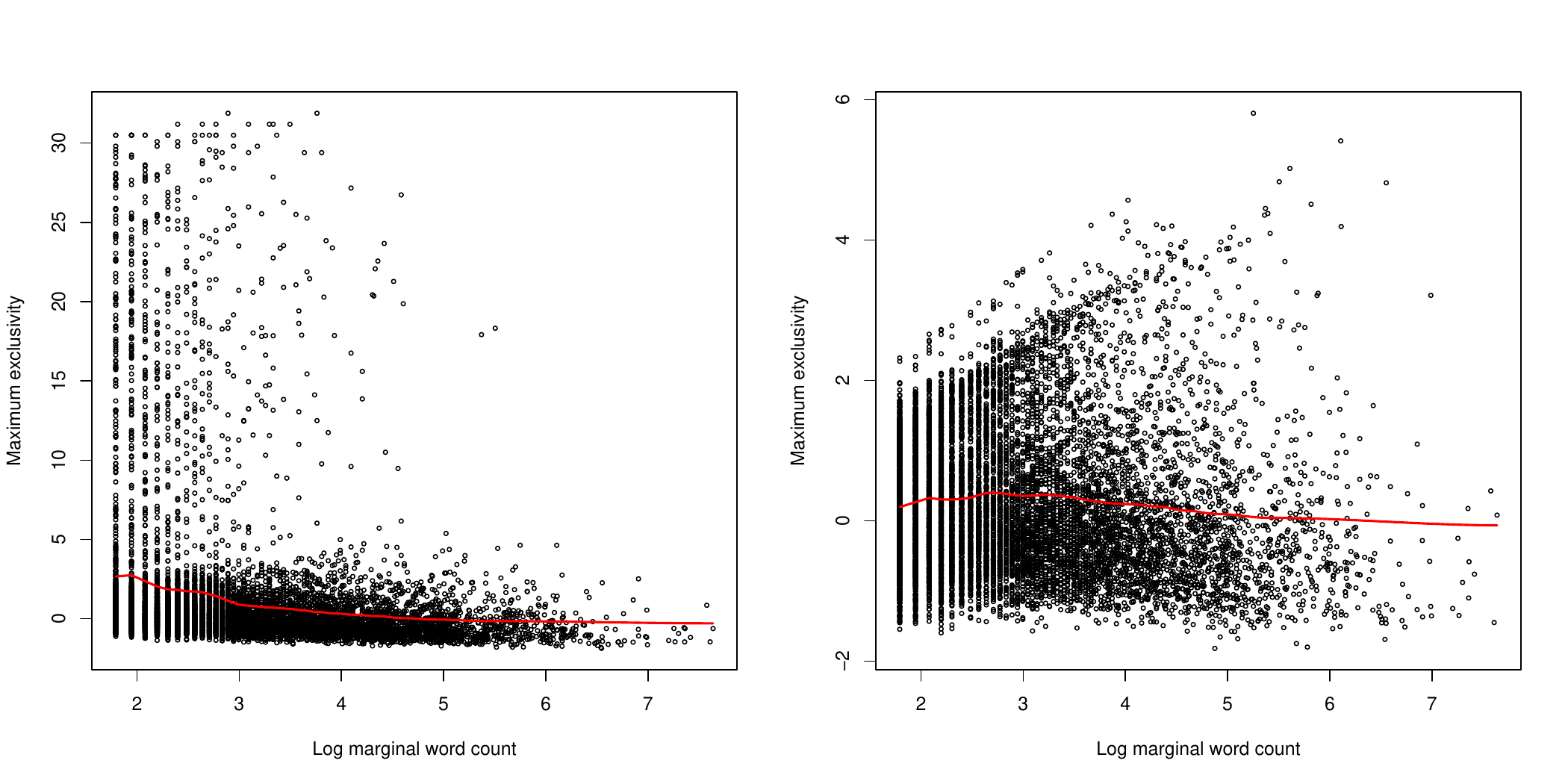}
\includegraphics[width=\textwidth]{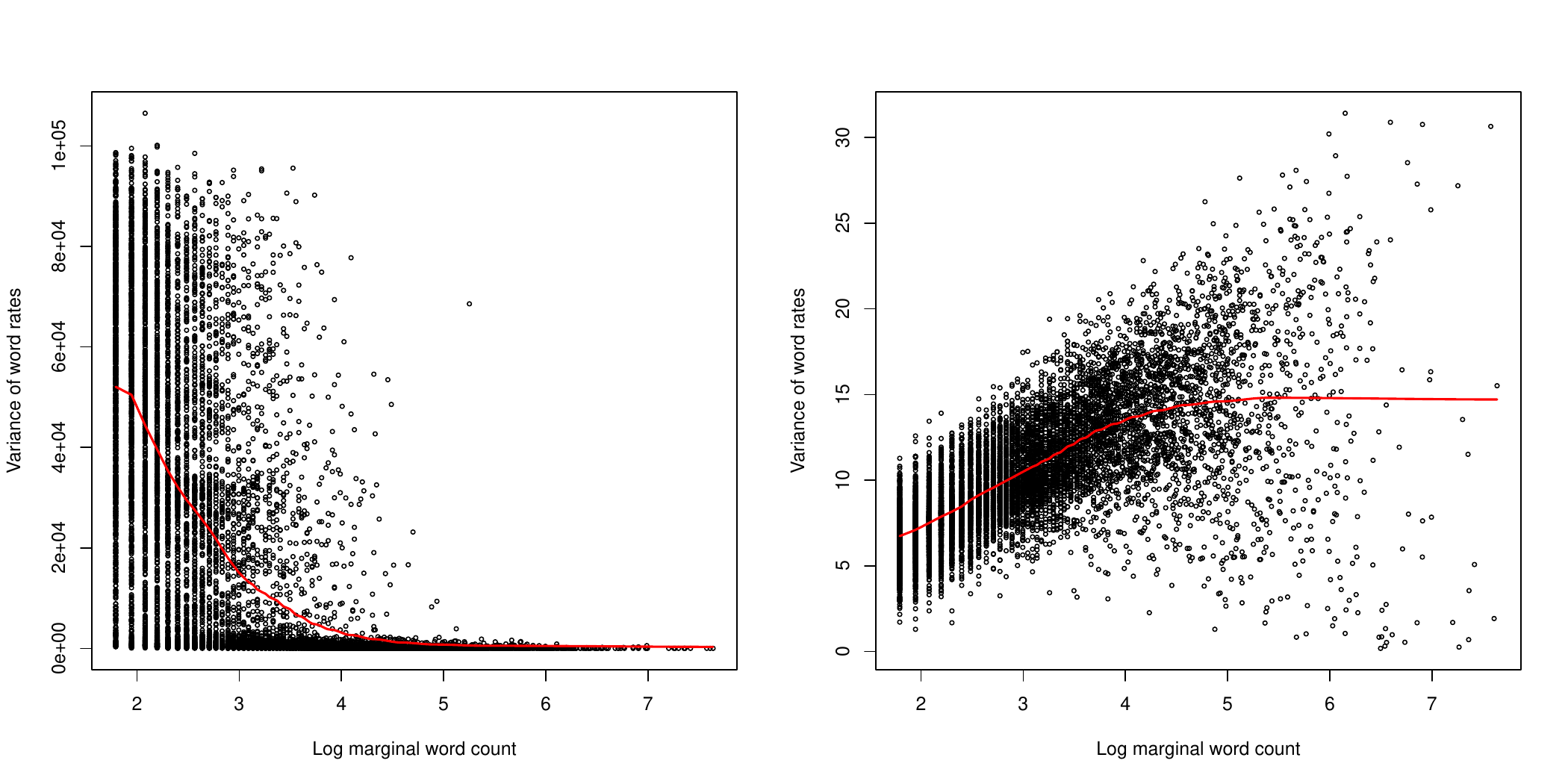}
\end{center}
\caption{Comparison of word/topic metrics for LDA (left panels) and the proposed model (right panels) fitted with 10 topics. The scatterplots show maximum exclusivity across topics (top panels), and  variance of word rates across topics (bottom panels). Constant loess smoother in red.}}
\label{ustext:fig:10comp}
\end{figure}
\begin{figure}[p]
\rev{
\begin{center}
Regularization within topic \hspace{90pt} Regularization across topics
\includegraphics[width=\textwidth]{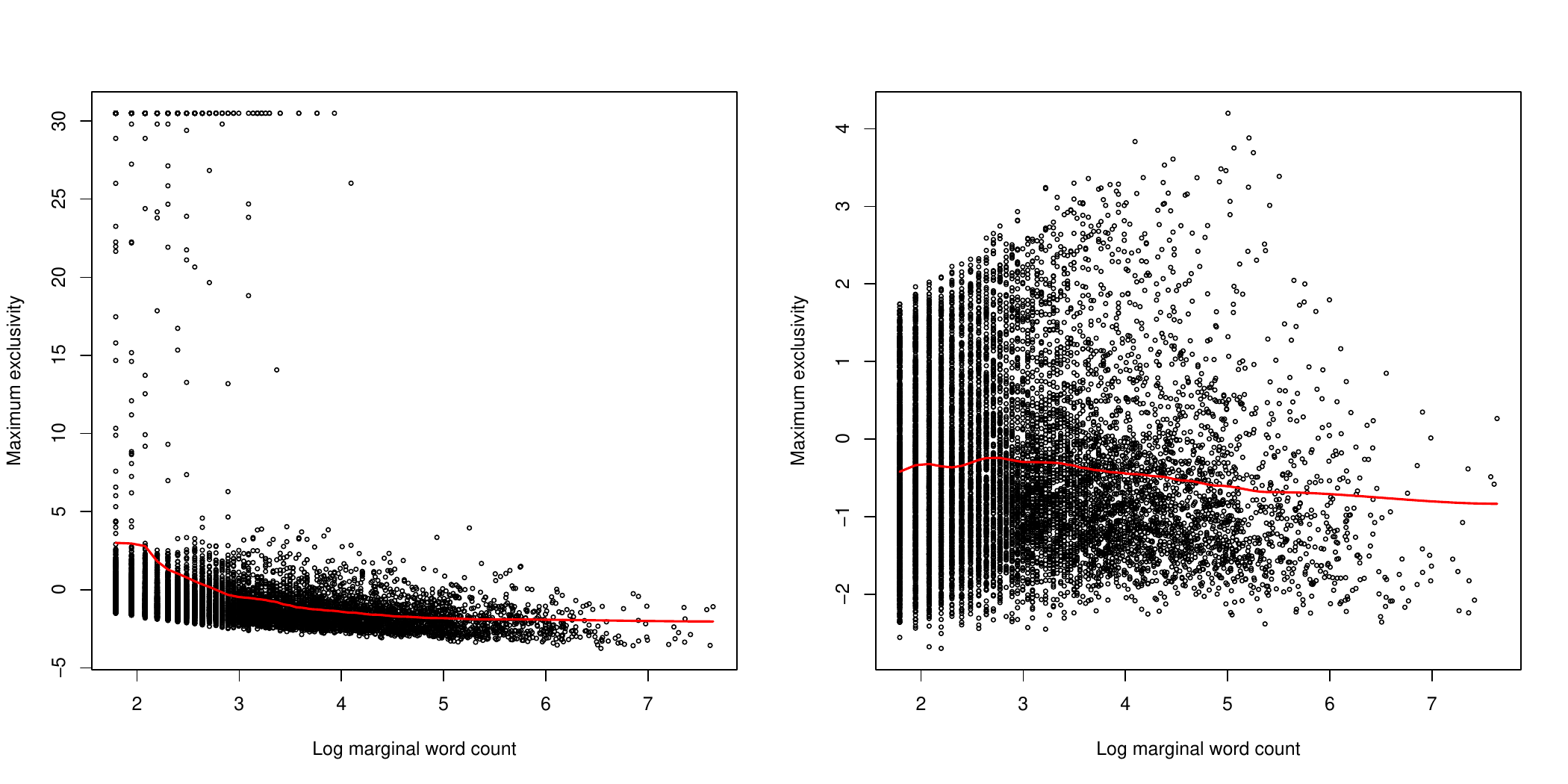}
\includegraphics[width=\textwidth]{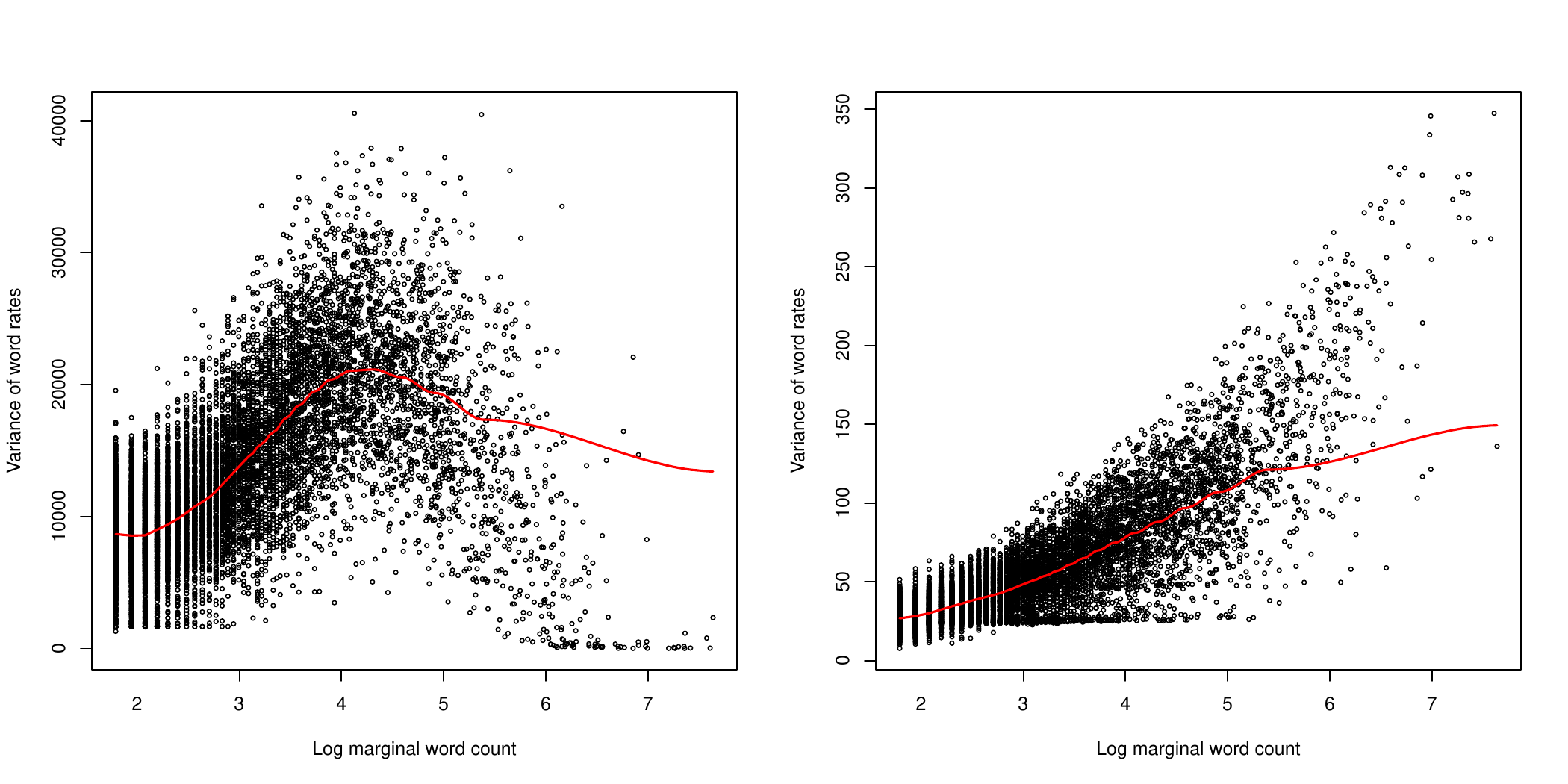}
\end{center}
\caption{Comparison of word/topic metrics for LDA (left panels) and the proposed model (right panels) fitted with 100 topics. The scatterplots show  maximum exclusivity across topics (top panels), and  variance of word rates across topics (bottom panels). Constant loess smoother in red.}}
\label{ustext:fig:100comp}
\end{figure}
The working hypothesis is that unregularized, or poorly regularized, estimates of exclusivity may promote rare words, which can produce word counts across topics that depart significantly from the uniform vector in a corpus even if their usage across topics is equal in expectation. As a result, exclusivity-based topic summaries might be dominated by rare words, regardless of their topical content. 

To gauge the severity of this problem, Figures \ref{ustext:fig:10comp} and \ref{ustext:fig:100comp} show scatterplots where the maximum exclusivity of each word across topics (top panels), and the variance of the estimated word rates (bottom panels) are plotted as a function of the marginal word count, for the LDA (left panels) and for proposed model (right panels). Figure \ref{ustext:fig:10comp} refers to model fits with 10 topics. Figure \ref{ustext:fig:100comp} refers to model fits with 100 topics.
%
The top panels show that LDA assigns its highest exclusivity scores to words with less than 100 total occurrences, whose scores dominate those of high frequency words by several orders of magnitude (on the logit scale). The bottom panels show that LDA assigns the highest variance of word rates across topics to words with less than 100 total occurrences. In contrast, the proposed model reverses these relationships in all these scatterplots, giving the highest maximum exclusivity and variance to the most frequent words. These patterns are consistent across model sizes.
The variance results are consistent with previous work showing that LDA leads to highly variable word rates across topics, especially for rare words \citep{eisensteinetal11}. 

}

\section{Concluding remarks}


\rev{The big idea that emerges from our work is the need to quantify how words are used differentially across topics as well as within them in order to summarize topical content in an interpretable fashion; we refer to these dimensions of content as word exclusivity and frequency. Topical summaries that focus on word frequency alone are often dominated by stop words or other terms used similarly across many topics.
Words can be visualized graphically in the exclusivity vs. frequency space, or these dimensions can be combined into a scalar quantity, such as the FREX score  proposed in Section \ref{sec:estimands}, to obtain a univariate measure of the topical content for words in each topic.}

\rev{Estimates of exclusivity based on rates of word occurrence regularized within a topic, as in LDA, are biased toward rare words due to sensitivity to small differences in estimated use across topics, as shown in Section \ref{sec:stability}. Topic models with regularization strategies borrowed from LDA cannot regularize differential use due to topic normalization of usage rates; its symmetric Dirichlet prior on topic distributions regularizes within, not between, topic usage. While topic-regularized models can capture many important facets of word usage, they are not optimal for the estimands used in our analysis of topical content.}

\rev{Related issues that affect the interpretability of the output of topic models are the treatment of stop words, and the presence of baseline, or nonsense, topics.
In any word summary of a topic, there is an issue of top versus bottom of the list; that is, words down the list are arguably not associated with any one topic. In the proposed approach, stop words get assigned a low FREX score in any one topic, thus their contribution to the summary becomes negligible. 
The proposed model reduces the importance of the stop words by design, using regularization induced by sensible priors. For instance, in Figure \ref{fe-plot} the stop words lie along a line where exclusivity is constant and frequency varies from high (for non-contextual stop words) to low (for corpus specific stop words). By contrast, in most models with an within-topic regularization \citep[a la][]{bleietal03} the emphasis on frequency exacerbates the issue of stop words, artificially promoting them, and leading to the appearance of nonsense topics. These issues are well known, and research efforts have proposed ways to mitigate the relevance of stop words and nonsense topics, either at the model level or as at pre- or post-processing stages \citep[e.g., see][]{wallachetal09,mimnoetal09}.
The Amazon Turk experiments in Section \ref{sec:mturk} show that the number of non-coherent topics, which can be taken as a proxy for nonsense topics, is reduced using our model-based estimates of frequency and exclusivity. This evidence supports the argument that our approach leads to a smaller number of nonsense topics. Topic summaries based on the FREX score are more interpretable.}

HPC breaks from standard topic models by modeling topic-specific word counts as unnormalized count variates whose rates can be regularized both within and across topics to compute word frequency and exclusivity. It was specifically designed to produce stable exclusivity estimates in human-annotated corpora by smoothing differential word usage according to a semantically intelligent distance metric: proximity on a known hierarchy. This supervised setting is an ideal test case for our framework and will be applicable to many high value corpora such as the \textit{ACM library}, \textit{IMS} publications, the \textit{New York Times} and \textit{Reuters}, which all have professional editors and authors and provide multiple annotations to a hierarchy of labels for each document. 

HPC offers a complex challenge for full Bayesian inference. To offer a flexible framework for regularization, it breaks from the simple Dirichlet-Multinomial conjugacy of traditional models. Specifically, HPC uses Poisson likelihoods whose rates are smoothed across a known topic hierarchy with a Gaussian diffusion and a novel mixed membership model where document label and topic membership parameters share a Gaussian prior. The membership model is the first to create an explicit link between the distribution of topic labels in a document and of the words that appear in a document and allow for multiple labels. However, the resulting inference is challenging since, conditional on word usage rates, the posterior of the membership parameters involves Poisson and Bernoulli likelihoods of differing dimensions constrained by a Gaussian prior.

We offer two methodological innovations to make inference tractable. First, we design our model with parameters that divide cleanly into two blocks (the tree and document parameters) whose members are conditionally independent given the other block, allowing for parallelized, scalable inference. However, these factorized distributions cannot be normalized analytically and are the same dimension as the number of topics (102 in the case of \textit{Reuters}). We therefore implement a Hamiltonian Monte Carlo conditional sampler that mixes efficiently through high dimensional spaces by leveraging the posterior gradient and Hessian information. This allows HPC to scale to large and complex topic hierarchies that would be intractable for Random Walk Metropolis samplers. 

One unresolved bottleneck in our inference strategy is that the MCMC sampler mixes slowly through the hyperparameter space of the documents---the $\bm{\eta}$ and $\lambda^{2}$ parameters that control the mean and sparsity of topic memberships and labels. This is due to a large fraction of missing information in our augmentation strategy \citep{mengrubin90}. Conditional on all the documents' topic affinity parameters $\{\bm{\xi}_{d}\}_{d=1}^{D}$, these hyperparameters index a normal distribution with $D$ observations; marginally, however, we have much less information about the exact loading of each topic onto each document. While we have been exploring more efficient data augmentation strategies such as Parameter Expansion \citep{liuwu99}, we have not found a workable alternative to augmenting the posterior with the entire set of $\{\bm{\xi}_{d}\}_{d=1}^{D}$ parameters.

\subsection{Toward semi-automated topic onthologies}


The HPC model can  be leveraged to semi-automate the construction of topic ontologies targeted to specific domains, for instance, when fit to comprehensive human-annotated corpora such as \emph{Wikipedia}, \emph{The New York Times}, \emph{Encyclopedia Britannica}, or databases such as \emph{JSTOR} and the \emph{ACM repository}.
By learning a probabilistic representation of high quality topics, HPC output can be used as a gold standard to aid and evaluate other learning methods \citep{Bakalov:2012fk}.

Targeted ontologies have been a key factor in monitoring scientific progress in biology \citep{Ashb:etal:2000,kane:goto:2000}.
A hierarchical ontology of topics would lead to new metrics for measuring progress in text analysis.
It would enable an evaluation of the semantic content of any collection of inferred topics, thus finally allowing for a {\it quantitative comparison} among the output of topic models. Current evaluations are qualitative, anecdotal and unsatisfactory; for instance, authors argue that lists of most frequent words describing an arbitrary selection of topics inferred by a new model make sense intuitively, or that they are better then lists obtained with other models.

In addition to model evaluation, a news-specific ontology could be used use as prior to inform the analysis of unstructured text, including Twitter feeds, Facebook wall posts, and blogs. Unsupervised topic models infer a latent topic space that may be oriented around unhelpful axes, such as authorship or geography. Using a human-created ontology as a prior could ensure that a useful topic space is discovered without being so dogmatic as to assume that unlabeled documents have the same latent structure as labeled examples.

\bibliographystyle{plainnat}

\begin{thebibliography}{41}
\providecommand{\natexlab}[1]{#1}
\providecommand{\url}[1]{\texttt{#1}}
\expandafter\ifx\csname urlstyle\endcsname\relax
  \providecommand{\doi}[1]{doi: #1}\else
  \providecommand{\doi}{doi: \begingroup \urlstyle{rm}\Url}\fi

\bibitem[Adams et~al.(2010)Adams, Ghahramani, and
  Jordan]{adams-ghahramani-jordan-2010}
R.~P. Adams, Z.~Ghahramani, and M.~I. Jordan.
\newblock Tree-structured stick breaking for hierarchical data.
\newblock In J.~Shawe-Taylor, R.~Zemel, J.~Lafferty, and C.~Williams, editors,
  \emph{Advances in Neural Information Processing (NIPS) 23}, 2010.

\bibitem[Airoldi et~al.(2006)Airoldi, Anderson, Fienberg, and
  Skinner]{Airo:Ande:Fien:Skin:2006}
E.~M. Airoldi, A.~G. Anderson, S.~E. Fienberg, and K.~K. Skinner.
\newblock Who wrote {R}onald {R}eagan's radio addresses?
\newblock \emph{Bayesian Analysis}, 1\penalty0 (2):\penalty0 289--320, 2006.

\bibitem[Airoldi et~al.(2007)Airoldi, Fienberg, and Xing]{Airo:Fien:Xing:2006}
E.~M. Airoldi, S.~E. Fienberg, and E.~P. Xing.
\newblock Mixed membership analysis of genome-wide expression
  studies---attribute data.
\newblock arXiv no. 0711.2520, July 2007.

\bibitem[Airoldi et~al.(2008)Airoldi, Blei, Fienberg, and Xing]{airoldietal08}
E.~M. Airoldi, D.~M. Blei, S.E. Fienberg, and E.P. Xing.
\newblock Mixed-membership stochastic blockmodels.
\newblock \emph{Journal of Machine Learning Research}, 9:\penalty0 1981--2014,
  2008.

\bibitem[Aletras and Stevenson(2013)]{Aletras2013}
N.~Aletras and M.~Stevenson.
\newblock {Evaluating topic coherence using distributional semantics}.
\newblock In \emph{IWCS}, number 2009, 2013.

\bibitem[Ashburner et~al.(2000)Ashburner, Ball, Blake, Botstein, Butler,
  Cherry, Davis, Dolinski, Dwight, Eppig, Harris, Hill, {Issel-Tarver},
  Kasarskis, Lewis, Matese, Richardson, Ringwald, Rubinand, and
  Sherlock]{Ashb:etal:2000}
M.~Ashburner, C.~A. Ball, J.~A. Blake, D.~Botstein, H.~Butler, J.~M. Cherry,
  A.~P. Davis, K.~Dolinski, S.~S. Dwight, J.~T. Eppig, M.~A. Harris, D.~P.
  Hill, L.~{Issel-Tarver}, A.~Kasarskis, S.~Lewis, J.~C. Matese, J.~E.
  Richardson, M.~Ringwald, G.~M. Rubinand, and G.~Sherlock.
\newblock Gene ontology: {T}ool for the unification of biology. {T}he gene
  ontology consortium.
\newblock \emph{Nature Genetics}, 25\penalty0 (1):\penalty0 25--29, 2000.

\bibitem[Bakalov et~al.(2012)Bakalov, McCallum, Wallach, and
  Mimno]{Bakalov:2012fk}
A.~Bakalov, A.~McCallum, H.~Wallach, and D.~Mimno.
\newblock Topic models for taxonomies.
\newblock In \emph{Proceedings of the 12th ACM/IEEE-CS Joint Conference on
  Digital Libraries}, 2012.

\bibitem[Blei.(2012)]{blei:2012}
D.~Blei.
\newblock Introduction to probabilistic topic models.
\newblock \emph{Communications of the {ACM}}, 2012.
\newblock In press.

\bibitem[Blei and McAuliffe(2007)]{bleimcauliffe07}
D.~Blei and J.~McAuliffe.
\newblock Supervised topic models.
\newblock volume~21. Neural Information Processing Systems, 2007.

\bibitem[Blei et~al.(2003{\natexlab{a}})Blei, Griffiths, Jordan, and
  Tenenbaum]{bleihier03}
D.~Blei, T.~Griffiths, M.~Jordan, and J.~Tenenbaum.
\newblock {Hierarchical topic models and the nested Chinese restaurant
  process}.
\newblock NIPS, 2003{\natexlab{a}}.

\bibitem[Blei et~al.(2003{\natexlab{b}})Blei, Ng, and Jordan]{bleietal03}
D.~Blei, A.~Ng, and M.~Jordan.
\newblock Latent dirichlet allocation.
\newblock \emph{Journal of Machine Learning Research}, 2003{\natexlab{b}}.

\bibitem[Breiman(2001)]{Brei:2001b}
L.~Breiman.
\newblock Statistical modeling: {T}he two cultures.
\newblock \emph{Statistical Science}, 16\penalty0 (3):\penalty0 199--231, 2001.

\bibitem[Buntine and Jakulin(2006)]{Buntin:2006fk}
W.~Buntine and A.~Jakulin.
\newblock Discrete components analysis.
\newblock In \emph{Subspace, Latent Structure and Feature Selection}, volume
  3940 of \emph{Lecture Notes in Computer Science}, pages 1--33. Springer,
  2006.

\bibitem[Canny(2004)]{canny04}
J.~Canny.
\newblock {GAP: A} factor model for discrete data.
\newblock In \emph{Proceedings of the 27th Annual International ACM SIGIR
  Conference on Research and Development in Information Retrieval}, 2004.

\bibitem[Chang et~al.(2009)Chang, Boyd-Graber, Gerrish, Wang, and
  Blei]{changetal09}
J.~Chang, J.~Boyd-Graber, S.~Gerrish, C.~Wang, and D.~Blei.
\newblock {Reading tea leaves: How humans interpret topic models}.
\newblock Neural Information Processing Systems, 2009.

\bibitem[Eisenstein et~al.(2011)Eisenstein, Ahmed, and Xing]{eisensteinetal11}
J.~Eisenstein, A.~Ahmed, and E.~P. Xing.
\newblock {Sparse Additive Generative Models of Text}.
\newblock ICML, 2011.

\bibitem[Harman(1992)]{harman92}
D.~Harman.
\newblock {Overview of the first text retrieval conference (TREC-1)}.
\newblock In \emph{{Proceedings of the First Text Retrieval Conference
  (TREC-1)}}, pages 1--20, 1992.

\bibitem[Hotelling(1936)]{hote:1936}
H.~Hotelling.
\newblock Relations between two sets of variants.
\newblock \emph{Biometrika}, 28:\penalty0 321--377, 1936.

\bibitem[Hu et~al.(2011)Hu, Boyd-Graber, and Satinoff]{huetal11}
Y.~Hu, J.~Boyd-Graber, and B.~Satinoff.
\newblock {Interactive Topic Modeling}.
\newblock Association for Computational Linguistics, 2011.

\bibitem[Jia et~al.(2014)Jia, Miratrix, Yu, Gawalt, {El Ghaoui}, Barnesmoore,
  and Clavier]{Jia:2014fk}
J.~Jia, L.~Miratrix, B.~Yu, B.~Gawalt, L.~{El Ghaoui}, L.~Barnesmoore, and
  S.~Clavier.
\newblock Concise comparative summaries {(CCS)} of large text corpora with a
  human experiment.
\newblock \emph{Annals of Applied Statistics}, 8\penalty0 (1):\penalty0
  499--529, 2014.

\bibitem[Jolliffe(1986)]{Joll:1986}
I.~T. Jolliffe.
\newblock \emph{Principal Component Analysis}.
\newblock Springer-Verlag, 1986.

\bibitem[Kanehisa and Goto(2000)]{kane:goto:2000}
M.~Kanehisa and S.~Goto.
\newblock {KEGG}: {K}yoto encyclopedia of genes and genomes.
\newblock \emph{Nucleic Acids Research}, 28\penalty0 (1):\penalty0 27--30,
  2000.

\bibitem[Lewis et~al.(2004)Lewis, Yang, Rose, and Li]{lewisetal04}
D.~D. Lewis, Y.~Yang, T.~G. Rose, and F.~Li.
\newblock {RCV1: A New Benchmark Collection for Text Categorization Research}.
\newblock \emph{Journal of Machine Learning Research}, 5:\penalty0 361--397,
  2004.

\bibitem[Liu and Wu(1999)]{liuwu99}
J.~S. Liu and Y.~N. Wu.
\newblock Parameter expansion for data augmentation.
\newblock \emph{Journal of the American Statistical Association}, 94:\penalty0
  1264--1274, 1999.

\bibitem[McCallum et~al.(1998)McCallum, Rosenfeld, Mitchell, and
  Ng]{mccallumetal98}
A.~McCallum, R.~Rosenfeld, T.~Mitchell, and A.~Ng.
\newblock Improving text classification by shrinkage in a hierarchy of classes.
\newblock International Conference on Machine Learning, 1998.

\bibitem[McLachlan and Peel(2000)]{mclachlanpeel2000}
G.~McLachlan and D.~Peel.
\newblock \emph{Finite Mixture Models}.
\newblock Wiley, 2000.

\bibitem[Meng and Rubin(1991)]{mengrubin90}
{X.-L.} Meng and D.~B. Rubin.
\newblock Using em to obtain asymptotic variance-covariance matrices: The sem
  algorithm.
\newblock \emph{Journal of the American Statistical Association}, 86:\penalty0
  899--909, 1991.

\bibitem[Mimno et~al.(2007)Mimno, Li, and McCallum]{mimnoetal07}
D.~Mimno, W.~Li, and A.~McCallum.
\newblock Mixtures of hierarchical topics with pachinko allocation.
\newblock ICML, 2007.

\bibitem[Mimno et~al.(2011)Mimno, Wallach, Talley, Leenders, and
  McCallum]{mimnoetal09}
D.~Mimno, H.~Wallach, E.~Talley, M.~Leenders, and A.~McCallum.
\newblock {Optimizing Semantic Coherence in Topic Models}.
\newblock EMNLP, 2011.

\bibitem[Mosteller and Wallace(1964)]{Most:Wall:1964}
F.~Mosteller and D.L. Wallace.
\newblock \emph{Inference and {D}isputed {A}uthorship: {T}he {F}ederalist}.
\newblock Addison-Wesley, 1964.

\bibitem[Mosteller and Wallace(1984)]{Most:Wall:1984}
F.~Mosteller and D.L. Wallace.
\newblock \emph{Applied Bayesian and Classical Inference: The Case of ``The
  Federalist" Papers}.
\newblock Springer-Verlag, 1984.

\bibitem[Neal(2011)]{neal2011}
R.~Neal.
\newblock {MCMC using Hamiltonian dynamics}.
\newblock In Steve Brooks, Andrew Gelman, Galin~L. Jones, and Xiao-Li Meng,
  editors, \emph{Handbook of Markov Chain Monte Carlo}. Chapman \& Hall / CRC
  Press, 2011.

\bibitem[Newman et~al.(2010)Newman, Lau, Grieser, and Baldwin]{Newman2010}
D.~Newman, J.~H. Lau, K.~Grieser, and T.~Baldwin.
\newblock {Automatic evaluation of topic coherence}.
\newblock In \emph{Human Language Technologies}, number June, pages 100--108,
  2010.

\bibitem[Perotte et~al.(2012)Perotte, Bartlett, Elhadad, and
  Wood]{perotteetal12}
A.~Perotte, N.~Bartlett, N.~Elhadad, and F.~Wood.
\newblock {Hierarchically Supervised Latent Dirichlet Allocation}.
\newblock NIPS, 2012.

\bibitem[Ramage et~al.(2009)Ramage, Hall, Nallapati, and
  Manning.]{ramageetal09}
D.~Ramage, D.~Hall, R.~Nallapati, and C.~D. Manning.
\newblock {Labeled LDA: A supervised topic model for credit attribution in
  multi-labeled corpora}.
\newblock EMNLP, 2009.

\bibitem[Rubin et~al.(2012)Rubin, Chambers, Smyth, and Steyvers]{rubinetal12}
T.~Rubin, A.~Chambers, P.~Smyth, and M.~Steyvers.
\newblock Statistical topic models for multi-label document classification.
\newblock \emph{Machine Learning}, 88, 2012.

\bibitem[Sandhaus(2008)]{Sandhaus:2008fk}
E.~Sandhaus.
\newblock {The New York Times annotated corpus}.
\newblock Linguistic Data Consortium, October 2008.
\newblock URL \url{http://catalog.ldc.upenn.edu/LDC2008T19}.

\bibitem[Sohn and Xing(2009)]{sohnxing2009}
{K.-A.} Sohn and E.~P. Xing.
\newblock A hierarchical dirichlet process mixture model for haplotype
  reconstruction from multi-population data.
\newblock \emph{Annals of Applied Statistics}, 3:\penalty0 791--821, 2009.

\bibitem[Wallach et~al.(2009)Wallach, Mimno, and McCallum]{wallachetal09}
H.~Wallach, D.~Mimno, and A.~McCallum.
\newblock {Rethinking LDA: Why priors matter}.
\newblock NIPS, 2009.

\bibitem[Zhu and Xing(2011)]{zhuxing11}
J.~Zhu and E.~P. Xing.
\newblock {Sparse Topical Coding}.
\newblock UAI, 2011.

\bibitem[Zhu et~al.(2012)Zhu, Ahmed, and Xing]{Zhu:2012fk}
J.~Zhu, A.~Ahmed, and E.~P. Xing.
\newblock Medlda: {M}aximum margin supervised topic models.
\newblock \emph{Journal of Machine Learning Research}, 13:\penalty0 2237--2278,
  2012.

\end{thebibliography}

\newpage
\appendix
\section{Appendix: Implementing the parallelized HMC sampler}\label{sec:appendix_gibbs}

\subsection{Hamiltonian Monte Carlo conditional updates}
Hamiltonian Monte Carlo (HMC) is the key tool that makes high-dimensional, non-conjugate updates tractable for our Gibbs sampler. It works well for log densities that are unimodal and have relatively constant curvature. We outline our customized implementation of the algorithm here; a general introduction can be found in \cite{neal2011}.

HMC is a version of the Metropolis-Hastings algorithm that replaces the common Multivariate Normal proposal distribution with a distribution based on Hamiltonian dynamics. 
It can be used to make joint proposals on the entire parameter space or, as in this paper, to make proposals along the conditional posteriors as part of a Gibbs scan. While it requires closed form calculation of the posterior gradient and curvature to perform well, the algorithm can produce uncorrelated or negatively correlated draws from the target distribution that are almost always accepted.

A consequence of classical mechanics, Hamiltonian's equations can be used to model the movement of a particle along a frictionless surface. The total energy of the particle is the sum of its potential energy (the height of the surface relative to the minimum at the current position) and its kinetic energy (the amount of work needed to accelerate the particle from rest to its current velocity). Since energy is preserved in a closed system, the particle can only convert potential energy to kinetic (or vice versa) as it moves along the surface. 

Imagine a ball placed high on the side of the parabola $f(q)=q^2$ at position $q=-2$. Starting out, it will have no kinetic energy but significant potential energy due to its position. As it rolls down the parabola toward zero, it speeds up (gaining kinetic energy), but loses potential energy to compensate as it moves to a lower position. At the bottom of the parabola the ball has only kinetic energy, which it then translates back into potential energy by rolling up the other side until its kinetic energy is exhausted. It will then roll back down the side it just climbed, completely reversing its trajectory until it returns to its original position. 

HMC uses Hamiltonian dynamics as a method to find a distant point in the parameter space with high probability of acceptance. Suppose we want to produce samples from $f(\bm{q})$, a possibly unnormalized density.
Since we want high probability regions to have the least potential energy, we parameterize the surface the particle moves along as $U(\bm{q})=-\log{f(\bm{q})}$, which is the height of the surface and the potential energy of the particle at any position $\bm{q}$. The total energy of the particle, $H(\bm{p},\bm{q})$, is the sum of its kinetic energy, $K(\bm{p})$, and its potential energy, $U(\bm{q})$, where $\bm{p}$ is its momentum along each coordinate. After drawing an initial momentum for the particle (typically chosen as $\bm{p}\sim\mathcal{N}(\bm{0},\bm{M})$, where $\bm{M}$ is called the \textit{mass matrix}), we allow the system to evolve for a period of time---not so little that the there is negligible absolute movement, but not so much that the particle has time to roll back to where it started. 

HMC will not generate good proposals if the particle is not given enough momentum in each direction to efficiently explore the parameter space in a fixed window of time. The higher the curvature of the surface, the more energy the particle needs to move to a distant point. Therefore the performance of the algorithm depends on having a good estimate of the posterior curvature $\bm{\hat{H}}(\bm{q})$ and drawing $\bm{p}\sim\mathcal{N}(\bm{0},-\bm{\hat{H}}(\bm{q}))$. If the estimated curvature is accurate and relatively constant across the parameter space, the particle will have high initial momentum along directions where the posterior is concentrated and less along those where the posterior is more diffuse. 

Unless the (conditional) posterior is very well behaved, the Hessian should be calculated at the log-posterior mode to ensure positive definiteness. Maximization is generally an expensive operation, however, so it is not feasible to update the Hessian every iteration of the sampler. In contrast, the log-prior curvature is very easy to calculate and well behaved everywhere. This led us to develop the \textit{scheduled conditional HMC sampler} (SCHMC), an algorithm for nonconjugate Gibbs draws that updates the log-prior curvature at every iteration but only updates the log-likelihood curvature in a strategically chosen subset of iterations. We use this algorithm for all non-conjugate conditional draws in our Gibbs sampler.

Specifically, suppose we want to draw from the conditional distribution $p(\bm{\theta}|\bm{\psi}_{t},\bm{y})\varpropto{}p(\bm{y}|\bm{\theta},\bm{\psi}_{t})p(\bm{\theta}|\bm{\psi}_{t})$ in each Gibbs scan, where $\bm{\psi}$ is a vector of the remaining parameters and $\bm{y}$ is the observed data. Let $\mathcal{S}$ be the set of full Gibbs scans in which the log-likelihood Hessian information is updated (which always includes the first). For Gibbs scan $i\in\mathcal{S}$, we first calculate the conditional posterior mode and evaluate both the Hessian of the log-likelihood, $\log{}p(\bm{y}|\bm{\theta},\bm{\psi}_{t})$, and of the log-prior, $\log{}p(\bm{\theta}|\bm{\psi}_{t})$, at that mode, adding them together to get the log-posterior Hessian. We then get a conditional posterior draw with HMC using the negative Hessian as our mass matrix. For Gibbs scan $i\notin\mathcal{S}$, we evaluate the log-prior Hessian at the current location and add it our last evaluation of the log-likelihood Hessian to get the log-posterior Hessian. We then proceed as before. The SCHMC procedure is described in step-by-step detail in Algorithm \ref{hmc_algo}.

\begin{algorithm}
\caption{Scheduled conditional HMC sampler for iteration $i$}\label{hmc_algo}
\footnotesize
\SetKwData{Left}{left}\SetKwData{This}{this}\SetKwData{Up}{up}
\SetKwFunction{Union}{Union}\SetKwFunction{FindCompress}{FindCompress}
\SetKwInOut{Input}{input}\SetKwInOut{Output}{output}
\Input{$\bm{\theta}_{t-1}$, $\bm{\psi}_{t}$ (current value of other parameters), $\bm{y}$ (observed data), $L$ (number of leapfrog steps), $\epsilon$ (stepsize), and $\mathcal{S}$ (set of full Gibbs scans in which the likelihood Hessian is updated)}
\Output{$\bm{\theta}_{t}$}
\BlankLine
\vspace{0.5em}
$\bm{\theta}_{0}^{*}\leftarrow\bm{\theta}_{t-1}$\;
\vspace{1em}
\tcc{Update conditional likelihood Hessian if iteration in schedule}
\If{$i\in\mathcal{S}$}{
\vspace{0.5em}
$\hat{\bm{\theta}}\leftarrow\text{arg}\max_{\bm{\theta}}\left\{\log{}p(\bm{y}|\bm{\theta},\bm{\psi}_{t}) + \log{}p(\bm{\theta}|\bm{\psi}_{t})\right\}$\;
\vspace{0.5em}
$\hat{\bm{H}}_{l}(\bm{\theta})\leftarrow\frac{\partial^{2}}{\partial\bm{\theta}\partial\bm{\theta}^{T}}\left[\log{}p(\bm{y}|\hat{\bm{\theta}},\bm{\psi}_{t})\right]|_{\bm{\theta}=\hat{\bm{\theta}}}$\;
}
\vspace{1em}
\tcc{Calculate prior Hessian and set up mass matrix}
$\hat{\bm{H}}_{p}(\bm{\theta})\leftarrow\frac{\partial^{2}}{\partial\bm{\theta}\partial\bm{\theta}^{T}}\left[\log{}p(\bm{\theta}|\bm{\psi}_{t})\right]|_{\bm{\theta}=\bm{\theta}_{0}^{*}}$\;
\vspace{0.5em}
$\hat{\bm{H}}(\bm{\theta})\leftarrow\hat{\bm{H}}_{l}(\bm{\theta}) + \hat{\bm{H}}_{p}(\bm{\theta})$\;
\vspace{0.5em}
$\bm{M}\leftarrow-\hat{\bm{H}}(\bm{\theta})$\;
\vspace{1em}
\tcc{Draw initial momentum}
Draw $\bm{p}^{*}_{0}\sim\mathcal{N}(\bm{0},\bm{M})$\;
\vspace{1em}
\tcc{Leapfrog steps to get HMC proposal}
\For{$l\leftarrow 1$ \KwTo $L$}{
\vspace{0.5em}
$\bm{g}_{1}\leftarrow-\frac{\partial}{\partial\bm{\theta}}\left[\log{}p(\bm{\theta}|\bm{\psi}_{t},\bm{y})\right]|_{\bm{\theta}=\bm{\theta}_{l-1}^{*}}$\;
\vspace{0.5em}
$\bm{p}^{*}_{l,1}\leftarrow\bm{p}^{*}_{l-1} - \frac{\epsilon}{2}\bm{g}_{1}$\;
\vspace{0.5em}
$\bm{\theta}^{*}_{l}\leftarrow\bm{\theta}^{*}_{l-1} + \epsilon(\bm{M}^{-1})^{T}\bm{p}^{*}_{l,1}$\;
\vspace{0.5em}
$\bm{g}_{2}\leftarrow-\frac{\partial}{\partial\bm{\theta}}\left[\log{}p(\bm{\theta}|\bm{\psi}_{t},\bm{y})\right]|_{\bm{\theta}=\bm{\theta}_{l}^{*}}$\;
\vspace{0.5em}
$\bm{p}^{*}_{l}\leftarrow\bm{p}^{*}_{l,1} - \frac{\epsilon}{2}\bm{g}_{2}$\;
}
\vspace{1em}
\tcc{Calculate Hamiltonian (total energy) of initial position}
$K_{t-1}\leftarrow\frac{1}{2}(\bm{p}^{*}_{0})^{T}\bm{M}^{-1}\bm{p}^{*}_{0}$\;
\vspace{0.5em}
$U_{t-1}\leftarrow-\log{}p(\bm{\theta}_{0}^{*}|\bm{\psi}_{t},\bm{y})$\;
\vspace{0.5em}
$H_{t-1}\leftarrow{}K_{t-1} + U_{t-1}$\;
\vspace{1em}
\tcc{Calculate Hamiltonian (total energy) of candidate position}
$K^{*}\leftarrow\frac{1}{2}(\bm{p}^{*}_{L})^{T}\bm{M}^{-1}\bm{p}^{*}_{L}$\;
\vspace{0.5em}
$U^{*}\leftarrow-\log{}p(\bm{\theta}_{L}^{*}|\bm{\psi}_{t},\bm{y})$\;
\vspace{0.5em}
$H^{*}\leftarrow{}K^{*} + U^{*}$\;
\vspace{1em}
\tcc{Metropolis correction to determine if proposal accepted}
Draw $u\sim\text{Unif}[0,1]$\;
\vspace{0.5em}
$\log{}r\leftarrow{}H_{t-1}-H^{*}$\;
\vspace{0.5em}
\eIf{$\log{}u<\log{}r$}{$\bm{\theta}_{t}\leftarrow{}\bm{\theta}_{L}^{*}$}{$\bm{\theta}_{t}\leftarrow{}\bm{\theta}_{t-1}$}
\end{algorithm}

\subsection{SCHMC implementation details for HPC model}
In the previous section we described our general procedure for obtaining samples from unnormalized conditional posteriors, the SCHMC algorithm. In this section, we provide the gradient and Hessian calculations necessary to implement this procedure for the unnormalized conditional densities in the HPC model, as well as strategies to obtain the maximum of each conditional posterior.

\subsubsection{Conditional posterior of the rate parameters}

The log conditional posterior of the rate parameters for one word is:
\begin{multline*}
\log{}p(\bm{\mu}_{f}|\bm{W},\bm{I},\bm{l},\{\bm{\tau}_{f}^2\}_{f=1}^{V},\psi,\gamma^{2},\nu,\sigma^{2},\{\bm{\xi}_{d}\}_{d=1}^{D},\mathcal{T})\\
=\sum_{d=1}^{D}\log{}\textrm{Pois}(w_{fd}|l_{d}\bm{\theta}_{d}^{T}\bm{\beta}_{f})+ \log{}\mathcal{N}(\bm{\mu}_{f}|\psi\bm{1},\bm{\Lambda}(\gamma^{2},\bm{\tau}_{f}^2,\mathcal{T}))\\
=-\sum_{d=1}^{D}l_{d}\bm{\theta}_{d}^{T}\bm{\beta}_{f} + \sum_{d=1}^{D}w_{fd}\log{}(\bm{\theta}_{d}^{T}\bm{\beta}_{f}) - \frac{1}{2}(\bm{\mu}_{f}-\psi\bm{1})^{T}\bm{\Lambda}(\bm{\mu}_{f}-\psi\bm{1}).
\end{multline*}
Since the likelihood is a function of $\bm{\beta}_{f}$, we need to use the chain rule to get the gradient in $\bm{\mu}_{f}$ space:
\begin{multline*}
\frac{\partial}{\partial\bm{\mu}_{f}}\bigg[\log{}p(\bm{\mu}_{f}|\bm{W},\bm{I},\bm{l},\{\bm{\tau}_{f}^2\}_{f=1}^{V},\psi,\gamma^{2},\{\bm{\xi}_{d}\}_{d=1}^{D},\mathcal{T})\bigg]\\
= \frac{\partial{}l(\bm{\beta}_{f})}{\partial\bm{\beta}_{f}}\frac{\partial\bm{\beta}_{f}}{\partial\bm{\mu}_{f}} + \frac{\partial}{\partial\bm{\mu}_{f}}\bigg[\log{}p(\bm{\mu}_{f}|\{\bm{\tau}_{f}^2\}_{f=1}^{V},\psi,\gamma^{2},\mathcal{T})\bigg]\\
= -\sum_{d=1}^{D}l_{d}(\bm{\theta}_{d}^{T}\circ\bm{\beta}_{f}^{T}) + \sum_{d=1}^{D}\bigg(\frac{w_{fd}}{\bm{\theta}_{d}^{T}\bm{\beta}_{f}}\bigg)(\bm{\theta}_{d}^{T}\circ\bm{\beta}_{f}^{T}) - \bm{\Lambda}(\bm{\mu}_{f}-\psi\bm{1}),
\end{multline*}
where $\circ$ is the Hadamard (entrywise) product. The Hessian matrix follows a similar pattern:
\begin{multline*}
\bm{H}(\log{}p(\bm{\mu}_{f}|\bm{W},\bm{I},\bm{l},\{\bm{\tau}_{f}^2\}_{f=1}^{V},\psi,\gamma^{2},\{\bm{\xi}_{d}\}_{d=1}^{D},\mathcal{T})) = -\bm{\Theta}^{T}\bm{W}\bm{\Theta}\circ\bm{\beta}_{f}\bm{\beta}_{f}^{T} + \bm{G} - \bm{\Lambda},
\end{multline*}
where
\begin{equation*}
\bm{W} = \text{diag}\bigg(\bigg\{\frac{w_{fd}}{(\bm{\theta}_{d}^{T}\bm{\beta}_{f})^{2}}\bigg\}_{d=1}^{D}\bigg)
\end{equation*}
and
\begin{equation*}
\bm{G} = \text{diag}\bigg(\frac{\partial{}l(\bm{\beta}_{f})}{\partial\bm{\beta}_{f}}\circ\bm{\beta}_{f}^{T}\bigg) = \text{diag}\bigg(\frac{\partial{}l(\bm{\beta}_{f})}{\partial\bm{\mu}_{f}}\bigg).
\end{equation*}

We use the BFGS algorithm with the analytical gradient derived above to maximize this density for iterations where the likelihood Hessian is updated; this quasi-Newton method works well since the conditional posterior is unimodal. The Hessian of the likelihood in $\bm{\beta}$ space is clearly negative definite everywhere since $\bm{\Theta}^{T}\bm{W}\bm{\Theta}$ is a positive definite matrix. The prior Hessian $\bm{\Lambda}$ is also positive definite by definition since it is the precision matrix of a Gaussian variate. However, the contribution of the chain rule term $\bm{G}$ can cause the Hessian to become indefinite away from the mode in $\bm{\mu}$ space if any of the gradient entries are sufficiently large and positive. Note, however, that the conditional posterior is still unimodal since the logarithm is a monotone transformation.

\subsubsection{Conditional posterior of the topic affinity parameters}

The log conditional posterior for the topic affinity parameters for one document is:
\begin{multline*}
\log{}p(\bm{\xi}_{d}|\bm{W},\bm{I},\bm{l},\{\bm{\mu}_{f},\bm{\tau}_{f}^2\}_{f=1}^{V},\bm{\eta},\bm{\Sigma})\\
=l_{d}\sum_{f=1}^{V}\log{}\textrm{Pois}(w_{fd}|\bm{\beta}_{f}^{T}\bm{\theta}_{d}) + \log{}\text{Bernoulli}(\bm{I}_{d}|\bm{\xi}_{d}) + \log{}\mathcal{N}(\bm{\xi}_{d}|\bm{\eta},\bm{\Sigma})\\
=-l_{d}\sum_{f=1}^{V}\bm{\beta}_{f}^{T}\bm{\theta}_{d} + \sum_{f=1}^{V}w_{fd}\log{}(\bm{\beta}_{f}^{T}\bm{\theta}_{d}) - \sum_{k=1}^{K}\log(1+\exp(-\xi_{dk}))\\
 - \sum_{k=1}^{K}(1-I_{dk})\xi_{dk}-\frac{1}{2}(\bm{\xi}_{d}-\bm{\eta})^{T}\bm{\Sigma}^{-1}(\bm{\xi}_{d}-\bm{\eta}).
\end{multline*}

Since the likelihood of the word counts is a function of $\bm{\theta}_{d}$, we need to use the chain rule to get the gradient of the likelihood in $\bm{\xi}_{d}$ space. This mapping is more complicated than in the case of the $\bm{\mu}_{f}$ parameters since each $\xi_{dk}$ is a function of all elements of $\bm{\theta}_{d}$:
\begin{equation*}
 \nabla{l_{d}(\bm{\xi}_{d})}=\nabla{l_{d}(\bm{\theta}_{d})}^{T}\bm{J}(\bm{\theta}_{d}\rightarrow\bm{\xi}_{d}),
\end{equation*}
where $\bm{J}(\bm{\theta}_{d}\rightarrow\bm{\xi}_{d})$ is the Jacobian of the transformation from 
$\bm{\theta}$ space to $\bm{\xi}$ space, a $K\times{}K$ symmetric matrix. Let $S=\sum_{l=1}^{K}\exp\xi_{dl}$. Then
\begin{equation*}
 \bm{J}(\bm{\theta}_{d}\rightarrow\bm{\xi}_{d}) = 
S^{-2}\begin{bmatrix}
S\exp\xi_{d1}-\exp2\xi_{d1} & \ldots
& -\exp(\xi_{dK} + \xi_{d1}) \\
-\exp(\xi_{d1} + \xi_{d2}) & \ldots
& -\exp(\xi_{dK} + \xi_{d2}) \\
\vdots & \ddots
& \vdots \\
-\exp(\xi_{d1} + \xi_{dK}) & \ldots
& S\exp\xi_{dK}-\exp2\xi_{dK}
\end{bmatrix}.
\end{equation*}
The gradient of the likelihood of the word counts in terms of $\bm{\theta}_{d}$ is
\begin{equation*}
 \nabla{l_{d}(\bm{\theta}_{d})}= -l_{d}\sum_{f=1}^{V}\bm{\beta}_{f}^{T} + \sum_{f=1}^{V}\frac{w_{fd}\bm{\beta}_{f}^{T}}{\bm{\beta}_{f}^{T}\bm{\theta}_{d}}.
\end{equation*}
Finally, to get the gradient of the full conditional posterior, we add the gradient of the likelihood of the labels and of the normal prior on the $\bm{\xi}_{d}$:
\begin{multline*}
\frac{\partial}{\partial{\bm{\xi}_{d}}}\bigg[\log{}p(\bm{\xi}_{d}|\bm{W},\bm{I},\bm{l},\{\bm{\mu}_{f}\}_{f=1}^{V},\bm{\eta},\bm{\Sigma})\bigg]\\
=\nabla{l_{d}(\bm{\theta}_{d})}^{T}\bm{J}(\bm{\theta}_{d}\rightarrow\bm{\xi}_{d}) + (\bm{1}+\exp\bm{\xi}_{d})^{-1} - (\bm{1}-\bm{I}_{d}) - \bm{\Sigma}^{-1}(\bm{\xi}_{d}-\bm{\eta}).
\end{multline*}

The Hessian matrix of the conditional posterior is a complicated tensor product that is not efficient to evaluate analytically. Instead, we compute a numerical Hessian using the analytic gradient presented above at minimal computational cost. 

We use the BFGS algorithm with the analytical gradient derived above to maximize this density for iterations where the likelihood Hessian is updated. We have not been able to show analytically that this conditional posterior is unimodal, but we have verified this graphically for several documents and have achieved achieved very high acceptance rates for our HMC proposals based on this Hessian calculation.

\subsubsection{Conditional posterior of the $\tau^{2}_{fk}$ hyperparameters}

The variance parameters $\tau^{2}_{fk}$ independently follow an identical Scaled Inverse-$\chi^{2}$ with convolution parameter $\nu$ and scale parameter $\sigma^{2}$, while their inverse follows a $\text{Gamma}(\kappa_{\tau}=\frac{\nu}{2},\lambda_{\tau}=\frac{2}{\nu\sigma^2})$ distribution. The log conditional posterior of these parameters is:
\begin{multline*}
\log{}p(\kappa_{\tau},\lambda_{\tau}|\{\bm{\tau}_{f}^2\}_{f=1}^{V},\mathcal{T}) = (\kappa_{\tau}-1)\sum_{f=1}^{V}\sum_{k\in\mathcal{P}}\log{(\tau^{2}_{fk})^{-1}} \\
 - |\mathcal{P}|V\kappa_{\tau}\log\lambda_{\tau} - |\mathcal{P}|V\log\Gamma(\kappa_{\tau}) - \frac{1}{\lambda_{\tau}}\sum_{f=1}^{V}\sum_{k\in\mathcal{P}}(\tau^{2}_{fk})^{-1},
\end{multline*}
where $\mathcal{P}(\mathcal{T})$ is the set of parent topics on the tree. If we allow $i\in\{1,\ldots,N=|\mathcal{P}|V\}$ to index all the $f,k$ pairs and $l(\kappa_{\tau},\lambda_{\tau})=p(\{\bm{\tau}_{f}^2\}_{f=1}^{V}|\kappa_{\tau},\lambda_{\tau},\mathcal{T})$, we can simplify this to
\begin{equation*}
l(\kappa_{\tau},\lambda_{\tau}) = (\kappa_{\tau}-1)\sum_{i=1}^{N}\log{\tau^{-2}_{i}} - N\kappa_{\tau}\log\lambda_{\tau} - N\log\Gamma(\kappa_{\tau}) - \frac{1}{\lambda_{\tau}}\sum_{i=1}^{N}\tau^{-2}_{i}.
\end{equation*}

We then transform this density onto the $(\log\kappa_{\tau},\log\lambda_{\tau})$ scale so that the parameters are unconstrained, a requirement for standard HMC implementation. Each draw of $(\log\kappa_{\tau},\log\lambda_{\tau})$ is then transformed back to the $(\nu,\sigma^{2})$ scale. To get the Hessian of the likelihood in log space, we calculate the derivatives of the likelihood in the original space and apply the chain rule:
\begin{multline*}
\bm{H}\bigg(l(\log\kappa_{\tau},\log\lambda_{\tau})\bigg)=\\
\begin{bmatrix}
\kappa_{\tau}\frac{\partial{l(\kappa_{\tau},\lambda_{\tau})}}{\partial{\kappa_{\tau}}} + (\kappa_{\tau})^{2}\frac{\partial^{2}{l(\kappa_{\tau},\lambda_{\tau})}}{\partial{(\kappa_{\tau})^{2}}} & \kappa_{\tau}\lambda_{\tau}\frac{\partial^{2}{l(\kappa_{\tau},\lambda_{\tau})}}{\partial{\kappa_{\tau}}\partial{\lambda_{\tau}}}\\
\kappa_{\tau}\lambda_{\tau}\frac{\partial^{2}{l(\kappa_{\tau},\lambda_{\tau})}}{\partial{\kappa_{\tau}}\partial{\lambda_{\tau}}} & \lambda_{\tau}\frac{\partial{l(\kappa_{\tau},\lambda_{\tau})}}{\partial{\lambda_{\tau}}} + (\lambda_{\tau})^{2}\frac{\partial^{2}{l(\kappa_{\tau},\lambda_{\tau})}}{\partial{(\lambda_{\tau})^{2}}}
\end{bmatrix},
\end{multline*}
where
\begin{equation*}
\nabla{l(\kappa_{\tau},\lambda_{\tau})} = 
\begin{bmatrix}
\sum_{i=1}^{N}\log{\tau^{-2}_{i}} - N\log\lambda_{\tau} - N\psi(\kappa_{\tau})\\
-\frac{N\kappa_{\tau}}{\lambda_{\tau}}+\frac{1}{(\lambda_{\tau})^{2}}\sum_{i=1}^{N}\tau^{-2}_{i}                               
\end{bmatrix}
\end{equation*}
and
\begin{equation*}
\bm{H}\bigg(l(\kappa_{\tau},\lambda_{\tau})\bigg)=
\begin{bmatrix}
- N\psi'(\kappa_{\tau}) & -\frac{N}{\lambda_{\tau}}\\
-\frac{N}{\lambda_{\tau}} & \frac{N\kappa_{\tau}}{(\lambda_{\tau})^{2}}-\frac{2}{(\lambda_{\tau})^{3}}\sum_{i=1}^{N}\tau^{-2}_{i}
\end{bmatrix}.
\end{equation*}

Following Algorithm \ref{hmc_algo}, we evaluate the Hessian at the mode of this joint posterior. This is easiest to find on original scale following the properties of the Gamma distribution. The first order condition for $\lambda_{\tau}$ can be solved analytically:

\begin{equation*}
\lambda_{\tau,MLE}(\kappa_{\tau}) = \text{arg}\max_{\lambda_{\tau}}\bigg\{l(\kappa_{\tau},\lambda_{\tau})\bigg\} = \frac{1}{\kappa_{\tau}N}\sum_{i=1}^{N}\tau^{-2}_{i}.
\end{equation*}
\vspace{5pt}
We can then numerically maximize the profile likelihood of $\kappa_{\tau}$:
\vspace{5pt}
\begin{equation*}
\kappa_{\tau,MLE}=\text{arg}\max_{\kappa_{\tau}}\bigg\{l(\kappa_{\tau},\lambda_{\tau,MLE}(\kappa_{\tau}))\bigg\}.
\end{equation*}

The joint mode in the original space is then $(\kappa_{\tau,MLE},\lambda_{\tau,MLE}(\kappa_{\tau,MLE}))$. Due to the monotonicity of the logarithm function, the mode in the transformed space is simply $(\log\kappa_{\tau,MLE},\log\lambda_{\tau,MLE})$. We can be confident that the conditional posterior is unimodal: the Fisher information for a Gamma distribution is negative definite, and the log transformation to the unconstrained space is monotonic.

\end{document}